\documentclass{article}

\usepackage{microtype}
\usepackage{graphicx}
\usepackage{subfigure}
\usepackage{booktabs} 
\usepackage{subcaption}
\usepackage{hyperref}
\usepackage{comment}
\usepackage{enumitem}

\usepackage[accepted]{icml2025}

\usepackage{amsmath}
\usepackage{amssymb}
\usepackage{mathtools}
\usepackage{amsthm}

\usepackage[capitalize,noabbrev]{cleveref}

\theoremstyle{plain}

\theoremstyle{definition}

\theoremstyle{remark}

\usepackage[textsize=tiny]{todonotes}

\usepackage{xcolor}

\icmltitlerunning{Algorithmic Understanding of LLMs}

\begin{document}

\twocolumn[
\icmltitle{Position: We Need An Algorithmic Understanding of Generative AI \\
           }




\begin{icmlauthorlist}
\icmlauthor{Oliver Eberle}{1,2}
\icmlauthor{Thomas McGee}{3}
\icmlauthor{Hamza Giaffar}{4}
\icmlauthor{Taylor Webb}{5}
\icmlauthor{Ida Momennejad}{5}
\end{icmlauthorlist}

\icmlaffiliation{1}{Technische Universität Berlin, Berlin, Germany}
\icmlaffiliation{2}{BIFOLD--Berlin Institute for the Foundations of Learning and Data, Berlin, Germany}
\icmlaffiliation{3}{University of California Los Angeles, Los Angeles, USA}
\icmlaffiliation{4}{Halıcıoğlu Data Science Institute, University of California San Diego, San Diego, USA}
\icmlaffiliation{5}{Microsoft Research NYC, New York, USA}
\icmlcorrespondingauthor{Ida Momennejad}{idamo@microsoft.com}
\icmlcorrespondingauthor{Oliver Eberle}{oliver.eberle@tu-berlin.de}

\icmlkeywords{Machine Learning, ICML}

\vskip 0.3in
]



\printAffiliationsAndNotice{}

\begin{abstract}
What algorithms do LLMs actually learn and use to solve problems? Studies addressing this question are sparse, as research priorities are focused on improving performance through scale, leaving a theoretical and empirical gap in understanding emergent algorithms. This position paper proposes \textit{AlgEval}: a framework for systematic research into the algorithms that LLMs learn and use.
AlgEval aims to uncover algorithmic primitives, reflected in latent representations, attention, and inference-time compute, and their algorithmic composition to solve task-specific problems.
We highlight potential methodological paths and a case study toward this goal, focusing on emergent search algorithms. Our case study illustrates both the formation of top-down hypotheses about candidate algorithms, and bottom-up tests of these hypotheses via circuit-level analysis of attention patterns and hidden states. The rigorous, systematic evaluation of how LLMs actually solve tasks provides an alternative to resource-intensive scaling, reorienting the field toward a principled understanding of underlying computations.  Such \textit{algorithmic explanations} 
offer a pathway to human-understandable interpretability, enabling comprehension of the model's internal reasoning performance measures.
This can in turn lead to more sample-efficient methods for training and improving performance, as well as novel architectures for end-to-end and multi-agent systems. 
\end{abstract}
\section{Introduction}

\label{submission}

Large language models (LLMs) have soared to prominence, yet a fundamental question remains: What algorithms do LLMs actually use to solve problems? As the “gold rush” of scaling prioritizes practical breakthroughs, research priorities have centered on improving performance through scale, often regardless of guarantees or costs,  while interpretability efforts have largely focused on understanding isolated mechanisms. Algorithmic understanding is often left behind. \textbf{This position paper argues that the ML community should prioritize research on an \textit{algorithmic} understanding of generative AI}. 

Existing work on understanding algorithmic operations in LLMs \cite{zhou2024what, li2023transformersalgorithmsgeneralizationstability, vonoswald2024uncoveringmesaoptimizationalgorithmstransformers, yang2024interpretability}, while impressive, remain surprisingly few in number. Recent interpretability research has prioritized the exploratory analysis of low-level circuit mechanisms~\cite{olah2020zoom,olsson2022context}, often without clear hypotheses, and even position papers that note the importance of algorithmic understanding only mention it as one among many other directions \cite{pmlr-v235-vilas24a}. Algorithmic research on LLMs has taken a backseat among priorities, with theoretical work on the topic almost entirely lacking for both individual and multi-agent LLM systems. While multi-agent systems are now common solutions for reasoning and planning with Transformers \cite{wu2023autogen, webb2024LLMPFC, nisioti2024}, theoretical foundations for efficiently building them remain underexplored.

While scaling has led to impressive results on a wide range of tasks, its limits remain unclear. Scale, rather than hypothesis-driven methods, has become the prevailing drive of general-purpose architectures since the rise of deep learning in the 2010s. This \textit{Bitter Lesson} \cite{bitterlesson2019}, combined with the hypothesis that reward may be enough for the emergence of intelligence \cite{SILVER2021103535}, have led to an emphasis on data- and compute-heavy approaches, prioritizing training data and fine-tuning with existing architectures. This trend has also widened the gap between frontier models and interpretability research, severely limiting their transparency, trustworthiness, and compliance with AI regulations \cite{samek2021, 10.1145/3491209}. 

The field is increasingly encountering the limitations of available data and its quality, and the rising computational costs and diminishing returns of scale \cite{10.5555/3692070.3694094}. This is in contrast to biological intelligence and brains, which provide an existence proof for a far more data- and energy-efficient approach. Recent studies suggest that optimizing inference-time compute can be more beneficial than simply scaling parameters, prompting shifts from feedforward parameter growth to inference-time compute \citep{snell2024scaling}. Thus, given the environmental costs, scaling without understanding is not a sustainable path forward, particularly for multi-agent AI, where insights into system interactions are increasingly crucial. 

This position paper calls for prioritizing systematic research on the algorithmic understanding of generative AI. An algorithm is typically defined as a finite set of rules or operations for transforming inputs into outputs \cite{turing1936a, Knuth-1968}, which can be combined to form efficient strategies to solve larger problems. 
Algorithmic explanations of LLMs, therefore, involves uncovering the specific step-by-step procedures or “computational primitives" these models effectively learn and execute during task solving. 
By examining operations within a model’s architecture, its parameters, and inference process, we can comprehensively determine how the model arrives at its outputs.

A systematic framework for algorithmic understanding should address: a) What algorithms can generative AI learn, and how does this depend on factors such as model size, training data, fine-tuning, and in-context learning? b) Are there provable guarantees for any such algorithmic abilities? c) How can we build multi-agent systems in order to implement specific algorithms? d) How can we set algorithmic objectives for training and fine-tuning? e) How can we create a repository of algorithmic abilities? f) How can we study the selection and composition of these components to solve prompted tasks? g) How can we design architectures to guarantee specific algorithmic capacities? In what follows, we outline \textit{AlgEval}, a research program for algorithmic evaluation and understanding of generative AI.

\section{Related Work}
To capture the rich internal computations implemented by increasingly complex machine learning (ML) systems, efforts in explainable AI and mechanistic interpretability have shifted focus to the inner workings of generative models, introducing approaches to uncover internal circuits \cite{olah2020zoom, wang2023interpretability}, representations \cite{todd2024function}, dynamical motifs \cite{yang2024interpretability}, and computational subgraphs \cite{higherorder2022, pmlr-v162-geiger22a}, laying the foundation for an \textit{algorithmic understanding} of model predictions.

Interpretability research initially emerged for deep classification models, before the era of generative AI \cite{lipton2017mythosmodelinterpretability, MONTAVON20181}. Understanding classification models has largely focused on identifying relevant input features or heatmaps at intermediate layers, using methods such as perturbation-based \cite{NIPS2017_8a20a862}, attention-based \cite{abnar-zuidema-2020-quantifying}, and gradient-based \cite{baehrens2010explain, sundararajan2017axiomatic, transformerlrp2022} approaches. With the shift toward generative AI and sequence modeling, a principled understanding of internal processes, beyond input-output relations or isolated mechanisms, has become essential. This presents a significant technical challenge due to the scale and complexity of today's frontier models. Thus, some researchers have focused on smaller or synthetic language models for targeted analysis and empirical studies, e.g., GPT-2 small \cite{wang2023interpretability,acdc2024, hanna2024have} or toy Transformers \cite{liu2023transformerslearnshortcutsautomata, YXLA2024-gsm1}.

On the other hand, early efforts to analyze LLM mechanisms focused on localizing specific functions within isolated model components \cite{vig2019analyzingstructureattentiontransformer, clark-etal-2019-bert}, such as individual neurons \cite{gurnee2024languagemodelsrepresentspace,zhou2018interpreting, templeton2024scaling}, or individual attention heads \cite{mcdougall-etal-2024-copy}. More recent work has investigated how these components combine to form functional circuits \cite{olsson2022context, wang2023interpretability, tigges2024llmcircuitanalysesconsistent}, while other work has characterized the representations that support some higher-level computations, e.g., function vectors \cite{todd2024function}. Probing techniques have further been developed to assess whether specific properties can be accurately decoded from a model’s latent representations~\cite{conneau-etal-2018-cram, hewitt-manning-2019-structural}, particularly in the analysis of reasoning strategies~\cite{YXLA2024-gsm1}. While these approaches move toward a more integrated understanding, current findings are still largely fragmented, and we lack a solid theoretical foundation for understanding how these various components come together to implement algorithms.

\section{AlgEval: Toward Algorithmic Evaluation and Understanding of LLMs} 
\label{sec:algeval}

Algorithms consist of modular subroutines that exhibit compositionality, allowing them to be reused and recombined into efficient strategies for solving increasingly complex problems.
From this conceptual starting point, AlgEval proposes a path toward algorithmic evaluation and understanding of LLMs through algorithmic primitives and their composition, analogous to a vocabulary and grammar.
We then explore methods to evaluate them, from the common analysis of attention weights, latent representations, and circuit methods, to new approaches like inference-time compute and evaluating alternative solutions.

\begin{figure}[ht!]
\vskip -0.1in
\begin{center}
    \centering
    \includegraphics[width=1.\columnwidth]{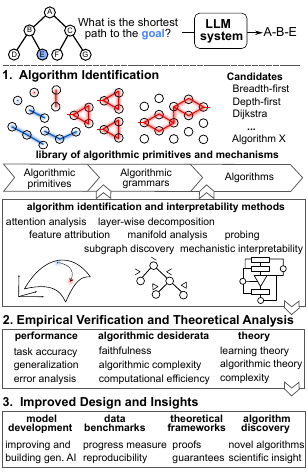}
    \vskip -0.1in
    \caption{AlgEval: A methodological path to prioritizing algorithmic evaluation and understanding of LLMs.}
    \label{fig:AlgEval_overview}
    \end{center}
    \vskip -0.3in
\end{figure}

A key challenge in the algorithmic understanding of LLMs is designing tasks that are complex enough to support specific algorithmic hypotheses and for which ground truth responses and strategies are available. An example is goal-directed navigation on deterministic graphs, Figure~\ref{fig:AlgEval_overview} (top). To solve such a task, classical search algorithms like breadth-first search (BFS), Dijkstra, or depth-first search (DFS) offer a verifiable algorithmic ground truth for evaluation. Moreover, heuristic, optimization-based approaches like simulated annealing \cite{kirkpatrick1983optimization, cerny1985thermodynamical, metropolis1953equation}, or amortized inference \cite{gershman2019amortized} could be at play. For instance, in amortized inference or “learning to infer,” rather than solving problems from scratch, the model compresses frequently used inference routines into a parametric function. In AlgEval, these hypotheses guide model analysis by evaluating internal mechanisms and inference-time generation to extract functional structure.

This hypothesis-driven approach re-centers the understanding and development of LLMs on scientific principles, emphasizing rigor in evaluating learned mechanisms. Drawing from Marr's three levels of analysis, i.e., computational, algorithmic, and implementation \cite{marr1982vision, pmlr-v235-vilas24a}, AlgEval prioritizes understanding mechanisms at the algorithmic level over their physical realization. By systematically testing algorithmic hypotheses, we also move beyond the assessment of behavioral goals at the computational level to uncover internal structures that enable robust problem-solving. Next, we describe what we refer to as \textit{algorithmic primitives}, and how models can piece them together to form algorithmic solutions. 

\subsection{Algorithmic Primitives and Vocabulary}

\textbf{Basic primitives.}  We define primitives to be the basic elements  necessary to realize a specific algorithm.  By iteratively breaking down an algorithm, a vocabulary of essential primitives can be obtained. In the context of LLMs, a variety of such primitives have been identified through methods from interpretability and data analysis. These approaches have led to the discovery of vector-based mechanisms, including function vectors for in-context learning mappings \cite{todd2024function, edelman2024the}, vector arithmetic computations \cite{merullo-etal-2024-language}, steering vectors \cite{yang2024interpretability}, copy suppression \cite{mcdougall-etal-2024-copy}, and key-value memory retrieval \cite{geva-etal-2021-transformer}. Additional primitives have been described that identify duplicate elements, and inhibit or increase attention to specific sequence elements \cite{wang2023interpretability}.

The functions of algorithmic primitives, as we defined them, should be predominantly domain-general and applicable to a wide range of sequence modification tasks. While the interpretability community provides a starting point to discover possibly domain general primitives, e.g., in-context learning (ICL) \cite{elhage2021mathematical, olsson2022context}, token binding \cite{vasileiou-2024-explaining, feng2024how}, or recognizing semantic relationships \cite{ren-etal-2024-identifying}, principled frameworks for connecting and combining these individual findings are mostly lacking. On the other hand, inspired by the Transformer architecture, a framework of programming primitives, termed  `restricted access sequence processing language' (RASP) \cite{pmlr-v139-weiss21a, zhou2024what}, has been developed to formalize the study of algorithmic implementations in Transformers' interpretable sequence operations, including basic select and aggregate manipulations. While helpful, RASP focuses on what could be built with LLMs in minimal scenarios, currently remaining inapplicable to understanding real world generative AI.

A key goal of AlgEval is the identification and evaluation of algorithmic primitives as clearly defined operations that can be assessed and validated at a lower level of complexity, and bridged to more sophisticated algorithmic levels.
Similar to current discussions on the emergence of universal representations \cite{pmlr-v235-huh24a}, this enables building a systematic catalog of universal primitives across models and tasks.

\textbf{Primitives as foundational algorithms.}
Essential primitives can be composed into domain-general basic algorithms for sequence generation tasks, leading to the discovery of algorithmic subgraphs and circuits that integrate multiple primitives and mechanisms. A number of disjointed findings are noteworthy. For instance, a network of induction, inhibition/excitation, duplicate token, and copy heads have been identified to solve the task of indirect object identification \cite{wang2023interpretability}.
Different algorithms have been discovered that solve modular arithmetic via combinations of circular embeddings and simple trigonometric operations \cite{nanda2023progress, zhong2023the}, which can be understood through different analytical solution approaches. By combining memory heads, that promote internally stored information,  with in-context heads, a mechanism for factual recall has been reconstructed \cite{yu-etal-2023-characterizing}. We consider these efforts as starting points for studying simple combinations of primitives. On the other hand, a set of theoretical studies focusing on complexity analysis \cite{elmoznino2024complexity}, algorithmic selection and assembly theory \cite{sharma2023assembly}, and combining and analyzing primitives via structured interactions  \cite{morris2019weisfeiler, interactions2022, higherorder2022, fumagalli2023shapiq}, call for an integration of theoretical and empirical contributions toward algorithmic discovery.

\textbf{Algorithms as primitives.} Recent studies have demonstrated how Transformers can display classic ML methods, including kernel-based approaches \cite{tsai-etal-2019-transformer}, support vector machines \cite{tarzanagh2023transformers}, Markov chains \cite{zekri2024largelanguagemodelsmarkov}, higher-order optimization methods for ICL \cite{fu2024transformers}, and temporal difference learning \cite{demircan2024tdsaes}. Further research is required to understand whether they can be understood as algorithmic primitives that can be combined, further broken down into more basic primitives, or both. This highlights the need to understand the extent to which primitives serve as task-specific or domain-general building blocks, and whether we can identify a systematic hierarchy of primitives.

\subsection{Algorithmic Composition}

\textbf{Compositionality.} The combination of algorithmic primitives into more complex algorithms is a form of \textit{compositionality} that, in principle, should support the construction of a vast number of combinations from a finite vocabulary. The search for algorithms is thus related to the broader debate about the extent to which LLMs are capable of compositional reasoning and generalization. While there is some evidence to support the existence of compositional representations and mechanisms in generative models~\citep{lepori2023break,campbell2024understanding}, there is also evidence for persistent failures in tasks that require compositional reasoning~\citep{lewis2022does,mitchell2023comparing,conwell2024relations}. Studying the algorithms used by these models can clarify whether their reasoning is truly compositional or reflects ineffective strategies like memorization~\cite{power2022grokking, qiu-etal-2022-evaluating}, or instabilities in representing rare events \cite{kandpal2023large}.

An algorithm's relevant primitives can be combined through several strategies. Previous works have explored optimization-based methods, such as learned graph structures and message passing \cite{pmlr-v162-geiger22a}, that enable dynamic interaction among components, while symbolic and compositional approaches provide explicit modular frameworks for task-specific solutions \cite{pmlr-v162-geiger22a, NEURIPS2023_f6a8b109}. ICL techniques, like providing grammar through input \cite{Galke2024}, can steer model behavior based on contextual cues. In the context of RASP, hard-coded aggregate functions can be used to nest several primitives, forming more complex functions, e.g., able to reverse or sort sequences \cite{pmlr-v139-weiss21a}. So far, it remains unclear whether  compositionality emerges when learning next-token prediction on, for example, a trillion tokens of compositional data, and to what extent it can be built into the architecture or training methods.

To foster deeper understanding of algorithmic composition, we need targeted empirical studies combined with building compositionality from first principles. Promising starting points include imposing known compositional constraints on the grammar, e.g., via hierarchical pyramids \cite{pyramids2017}, equivariant structure \cite{satorras2021n}, conditions imposed by the data-generating function \cite{wiedemer2023compositional}, or algorithmic complexity \cite{elmoznino2024complexity}. Assembly theory offers an evolutionary perspective on forming complex algorithmic grammars by selecting and recombining primitives  \cite{sharma2023assembly}.

\subsection{Methodologies for Algorithmic Evaluation} \label{sec:methodologies_algorithmic_evaluation}
AlgEval focuses on methodological advances necessary to identify, evaluate, and discover algorithms. This motivates combining existing interpretability techniques with novel approaches to capture the complexity of modern AI.

\textbf{Analyzing representations and attention.} Neural representations, patterns of activity across neural populations or intermediate Transformer layers, are increasingly recognized as key to understanding or modifying network computations \cite{zou2023representationengineeringtopdownapproach, sucholutsky2024gettingalignedrepresentationalalignment}. Analyses of representational similarities within or across layers, using an array of similarity measures sometimes inspired by cognitive science and neuroscience, elucidate how these layers transform information  \cite{kornblith2019similarity,  klabunde2024similarityneuralnetworkmodels, yousefi2024decodingincontextlearningneuroscienceinspired,sucholutsky2024gettingalignedrepresentationalalignment, Pro_NEURIPS2021_252a3dba, ENSD_pmlr-v238-giaffar24a}. For instance, LLM embeddings can reveal interpretable structures for deception detection by identifying a 2D subspace encoding true/false statements \cite{bürger2024truthuniversalrobustdetection}, and a three-stage process of deceptive behavior has been uncovered through low-dimensional projections \cite{yang2024interpretability}.

Complementary to representation analyses, attention in LLMs is commonly analyzed to identify message passing operations among tokens. Layer-wise attention scores help interpret token importance and internal model structures, including attention rollout and attention flow \cite{abnar-zuidema-2020-quantifying}.
To capture task-specific model processing, feature attribution methods compute feature importance scores, addressing the limits of attention analysis in providing faithful explanations \cite{wiegreffe-pinter-2019-attention}, with techniques like saliency \cite{9577970} and modified gradient methods \cite{transformerlrp2022,pmlr-v235-achtibat24a, jafari2024mambalrp}. Furthermore, internal analysis of attributions can aid in discovering relevant representational concepts \cite{kauffmann2022clustering, chormai2024disentangled}.

The integration of attention and representation analyses is a key feature of AlgEval, as information passing between tokens and representation analyses can uncover network structures and transformations, helping us understand algorithmic primitives and compositionality in problem-solving.
(see Section \ref{sec:cas_study}).

\textbf{Subgraphs and circuits.}  
The identification of relevant internal structure presents a current methodological frontier in our understanding of LLMs. 
Various methods have been proposed to extract circuits \cite{olah2020zoom, wang2023interpretability}, subgraphs \cite{higherorder2022, pmlr-v162-geiger22a}, feature interactions \cite{interactions2022, fumagalli2023shapiq, vasileiou-2024-explaining, kauffmann2024clever}, and causal symbolic models \cite{pmlr-v162-geiger22a}. Key techniques for discovery of internal structure include activation patching \cite{wang2023interpretability}, automatic circuit discovery \cite{acdc2024}, attribution patching \cite{syed-etal-2024-attribution, hanna2024have}, and graph explanations \cite{higherorder2022, sanford2024understanding}. Viewing  LLMs as graphs provides a complementary perspective of sequence processing as computations that extend across multi-hop neighborhoods that form substructures \cite{besta2024demystifyinghigherordergraphneural}, build motifs and compute higher-order interactions across neurons \cite{interactions2022,higherorder2022, fumagalli2023shapiq}.

\subsection{Identifying and Structuring Primitives}
\label{sec:identifying}
Methodologically, we have identified the following five steps as crucial components of AlgEval: (a) \textbf{Identify and form a library of primitives}  which can grow over time and anchor corresponding tasks and mechanisms. Each algorithmic primitive can correspond to multiple tasks and algorithms, supporting different mechanistic implementations. Primitives can be either hypothesis-based, rooted in decades of theoretical algorithm research, or empirically observed. (b) \textbf{Build a collection of simple tasks} which require a set of primitives for their solution. Examples include sequence induction~\cite{olsson2022context}, or copying a sequence of unique tokens~\cite{zhou2024what}. (c) \textbf{Create a library of mechanisms} that implement primitives, along with corresponding interpretability and analysis tools to identify them as discussed in Section~\ref{sec:methodologies_algorithmic_evaluation}.  (d) \textbf{Analysis of Composition} is crucial for identifying primitives, as it involves understanding how they combine, how algorithms are implemented across layers and inference, and whether compositional patterns generalize across tasks and models. (e) \textbf{Ablations} serve to identify  and evaluate primitives' role, which necessitates developing tools for causal intervention and ablation~\cite{pmlr-v162-geiger22a, pmlr-v236-talon24a}, as well as using statistical tests. Newly discovered primitives, along with their associated tasks, mechanisms, and methods, are then added to the growing library of primitives for future analyses and integration into new models.

\subsection{New Directions for Algorithmic Analysis}\label{inf_time_compute_section}

\textbf{The role of in-context learning.} Recent work has explored how ICL improves transformer performance \cite{fu2024transformers}, one proposing Transformers are algorithms \cite{li2023transformersalgorithmsgeneralizationstability}. 
A recent paper analyzed changes in attention and representation due to ICL and how it related to improvements in behavior \cite{yousefi2024decodingincontextlearningneuroscienceinspired}.
While that work did not focus on interpreting the effect of these changes on algorithmic primitives or composition, one important future direction will be to analyze the algorithmic consequences of ICL, such as promoting the use of specific algorithms.

\textbf{Inference-time compute.}
Inference-time compute has recently arisen as a new paradigm for reasoning with LLMs. In this approach, rather than solving a problem via a single feedforward pass, the autoregressive outputs of the model can be used to perform intermediate computations. Examples of this approach include chain-of-thought~\citep{wei2022chain}, explicit tree search~\citep{yao2024tree}, agent-based approaches~\citep{webb2024LLMPFC}, and models such as o1~\citep{jaech2024openai} and the open-source R1~\citep{deepseekai2025deepseekr1incentivizingreasoningcapability} that are trained to perform inference-time compute via amortized optimization. Some have even argued for scaling inference-time compute instead of scaling parameters or training~\citep{snell2024scaling}.

AlgEval also applies to emergent algorithms at inference time, where  sequential outputs can be more amenable to algorithmic analysis than high-dimensional feedforward computations. 
For example, LLMs can be trained to explicitly search via their outputs, implementing search procedures like exploration and backtracking through traces provided in context (see \textit{in-context search, SearchFormer, Stream of Search}) \cite{gandhi2024stream, lehnert2024beyond}.
This is particularly revealing when models acquire algorithms through amortized inference for downstream tasks rather than via supervised fine-tuning or ICL, potentially yielding novel or emergent solutions. 

While no existing work systematically examines inference-time emergent algorithms, likely due to their novelty, the core AlgEval principles of identifying algorithmic primitives, forming top-down hypotheses, and bottom-up testing remain applicable.
There may also be interactions between feedforward algorithms and inference-time compute, where certain processes like search are offloaded to the output space, allowing the feedforward pass to specialize in different primitives. Finally, it is crucial to ensure causal linkage between inference-time outputs and actual performance, since chain-of-thought can be unfaithful or unrelated to the model’s real reasoning \cite{turpin2024language,stechly2024chain}.

\textbf{Reinforcement learning and memory compression.}
A recent open-source LLM, DeepSeek R1 \cite{deepseekai2025deepseekr1incentivizingreasoningcapability}, used reinforcement learning (RL) and inference-time compute to match the performance of OpenAI's o1 at a fraction of the training cost. Interestingly, this model displayed an apparently emergent form of backtracking (referred to as an ‘aha moment’). A future direction of research is to study to what extent this behavior emerged purely as a consequence of training with RL, as opposed to relying on documents in the base model’s training data that include similar examples of backtracking (annotated math solutions). Given o1 was supposedly also trained on annotated math solutions, it is important to study how training with RL, particularly Group Relative Policy Optimization (GRPO) \cite{shao2024deepseekmathpushinglimitsmathematical}, is necessary to elicit this behavior. Open-source models like DeepSeek R1 offer an opportunity to study how training data and RL shape their  algorithmic vocabulary and compositions thereof. A key question herein is whether RL led to cached strategies for amortized inference, enabling ``learning to infer," by compressing prior inferences \cite{gershman2019amortized, radev2020amortizedbayesianinferencemodels}. Such strategies can link learning to compressed memory representations, which is also observed in the human brain and behavior \cite{Momennejad2017Successor, Russek2017Predictive, Momennejad2020LearningStructures, Brunec2021Predictive, momennejad_forthcoming}.

\section{Case Study} \label{sec:cas_study} 
\begin{figure}[t!]
    \centering\centerline{\includegraphics[width=\columnwidth]{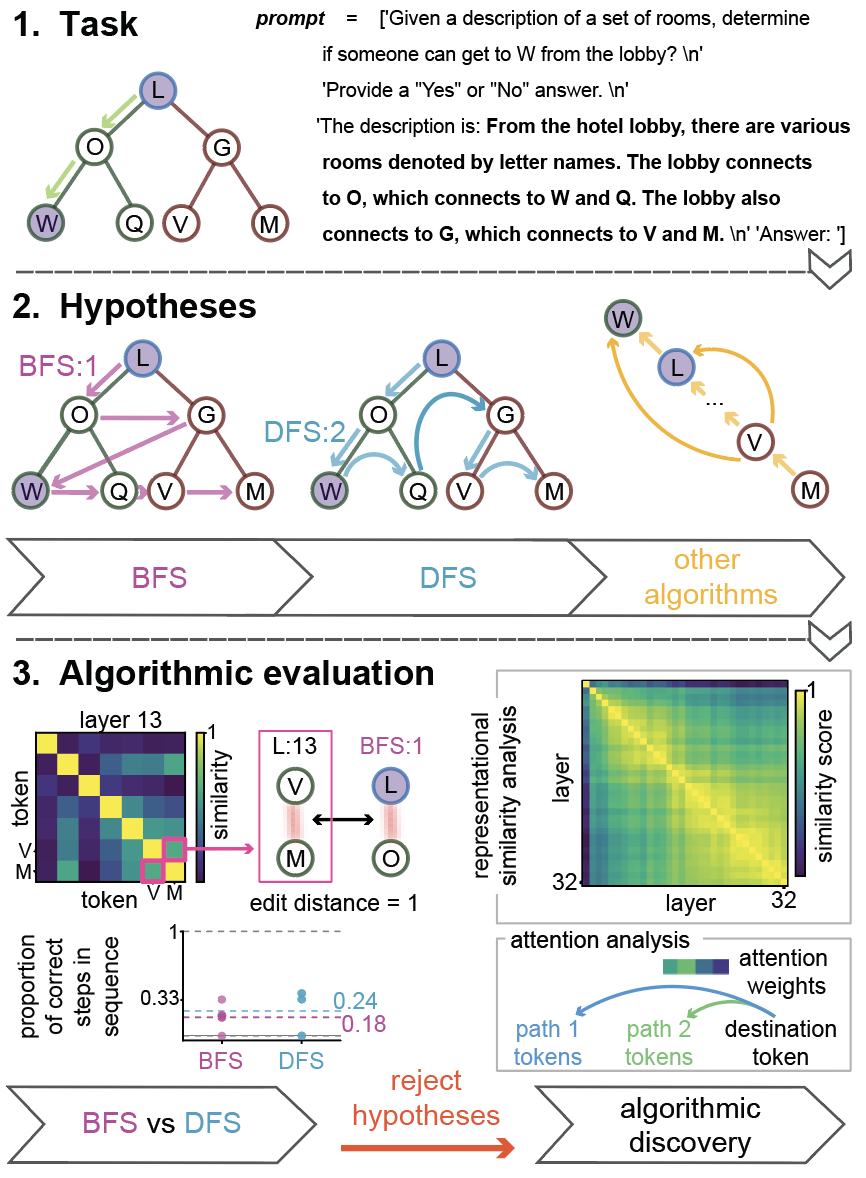}}
    \vskip -0.1in
    \caption{Case study. A graph navigation task, algorithmic hypotheses (possible rollouts for DFS and BFS are shown), and potential methods for algorithmic evaluation.}
    \label{fig:caseSchema}
    \vskip -0.2in
\end{figure}

To ground our position in an empirical example, we conducted a case study focused on LLMs, which have been shown to perform poorly on graph navigation and multi-step planning tasks \cite{momennejad2023cogeval}.
In cases where they do succeed, it remains unclear how they solve these problems, e.g., whether they implement classic search algorithms or use other strategies. To address this question, we studied the algorithms used by widely used LLMs, instruction-tuned Llama-3.1 with 8B and 70B parameters, in the context of graph navigation.
We considered a simple tree graph structure, presented in a prompt that describes how rooms (nodes) connect to one another (edges) and  tasks the model with determining whether a direct path from the start to the goal node exists (Figure~\ref{fig:caseSchema}).

A straightforward hypothesis to test is that the model may use standard search algorithms like DFS, BFS, or Dijkstra to find paths between graph nodes, with each layer potentially representing one search step.
This analysis assumes that each layer corresponds to one visited node or node pair and that multiple connections may be evaluated simultaneously. If the layers successfully identify the correct path, we can infer its sequence of visited nodes by examining which node tokens receive the highest attention or exhibit the strongest representational similarity across layers 
\cite{kriegeskorte2008representational, kornblith2019similarity,manvi2024adaptiveinferencetimecomputellms}. These combined analyses of attention and representation provide insight into the potential step-by-step algorithmic procedures underlying the LLM’s graph exploration.

\textbf{Prompt.} We introduce the model to a two-step tree graph following the prompt from~\citet{momennejad2023cogeval}, which demonstrated that LLMs struggle with graph navigation and especially tree search. The model is tasked with determining the validity of a given path, producing a single token output: `yes' or `no'. The full prompt and task for starting from the `lobby' and goal location W are shown in Figure~\ref{fig:results}a. Note that some nodes appear multiple times in the prompt (e.g., `lobby' is repeated four times), thus in our analysis in Figure~\ref{fig:results}c, we display the representation trajectory for each appearance of a node.

\subsection{Cascading Attention Analysis} 

Analyzing layer-wise attention matrices in LLMs \cite{vaswani2017attention} provides a direct and commonly used way to trace message-passing operations among tokens. 
To better understand how the model's algorithm leverages attention, we analyzed 1) the attention from each graph node to its preceding nodes and the goal, and 2) the attention from the final token to all nodes. The former allowed us to test how the graph nodes ``search over themselves" across layers, while the latter allowed us to test how the model searches over relevant tokens at inference time. We next present results on Llama-3.1-8B with additional analyses of the 70B model presented in Appendix~\ref{app:70b}.

\begin{figure*}[htbp]
    \centering
    \centerline{\includegraphics[width=\textwidth]{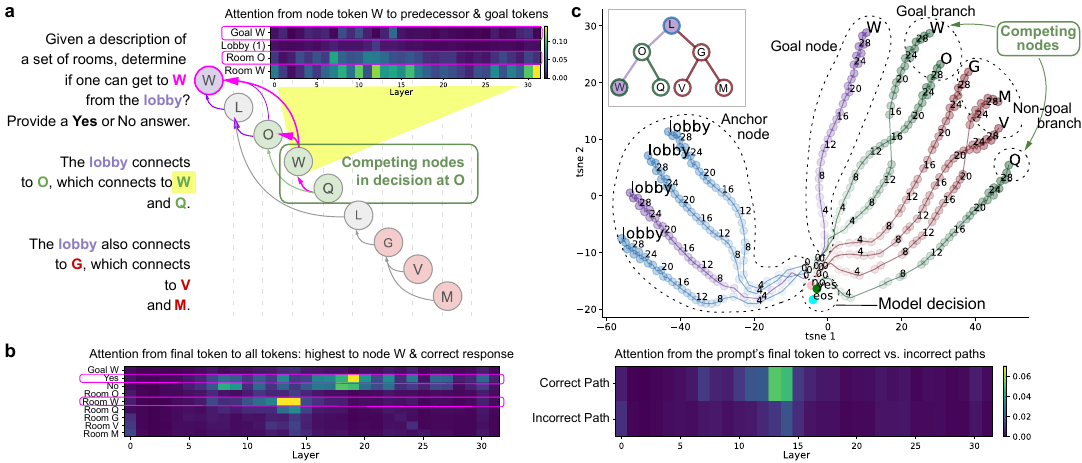}}
    \vskip -0.1in
    \caption{Case study on algorithmic discovery. (\textbf{a}) Attention heatmap: from goal to the correct vs. incorrect pathway. (\textbf{b}) Attention is mostly allocated to the goal location M, followed by V and G on the correct path. (\textbf{c}) Separation of room representations across layers.}
    \label{fig:results}
    \vskip -0.1in
\end{figure*}

To test whether the model performs search over graph nodes across layers, we analyzed how the final token's attention shifts between correct and incorrect paths as shown in Figure~\ref{fig:results}.
We first used a linear mixed-effects model with logit-transformed average attention as the outcome variable, pathway (correct or incorrect) as a fixed effect, and layer number as a random intercept. Results indicated that the final token allocated significantly more attention to rooms in the correct pathway than to those in the incorrect one (\textit{b} = 0.33,  \textit{SE} = 0.07, \textit{t}(2015) = 4.51, $\textit{p} < .001$) (Appendix Table \ref{app:tab:ttest_attn_correct}). The model included a random intercept for layer number (variance = 0.96, \textit{SD} = 0.98), and the residual variance was 2.80 (\textit{SD} = 1.67). A layer-wise analysis showed that the final token allocated significantly more attention to rooms on the correct path in 14 of 32 layers (Appendix Table \ref{app:tab:ttest_attn_correct_layers}), with only three layers exhibiting significantly more attention to the incorrect path (Appendix Table \ref{app:tab:ttest_attn_incorrect_layers}).

Analyzing attention to individual nodes, response tokens, and the goal token revealed an interpretable layer-by-layer sequence leading to the correct response: early-to-mid layers attended to pairwise node links, while attention to the goal node W peaked in layers 13–14, and the final token's attention shifted to the correct response around layer 19. 

The cascading attentional spread from each node to its predecessors may incrementally direct attention to the correct path, suggesting a \textbf{policy-dependent} algorithm rather than an exhaustive search like BFS or DFS. The model seems to (1) incrementally attend to the path leading to the goal via query-key attention weights, then (2) attend to the goal token in later layers as presented in Figure~\ref{fig:results}.

\subsection{Analysis of Feedforward Representations}

To characterize how graph representations contained in feedforward activity change across LLM layers, we first defined the token-by-token representational similarity matrices $V^i$ for each layer, indexed $i$, entries of which $V^i_{xy} = u_{i,x}^Tu_{i,y}$ are inner products between activation vectors $u_{i,x}$ and $u_{i,y}$, for room tokens $x$ and $y$ respectively.  To ask if we could identify discrete changes in the graph representational geometry between neighboring layers that might be interpreted as steps of an algorithm, we computed layer-by-layer similarity matrices, $S^d$ using similarity measure $d$  \cite{kornblith2019similarity, Pro_NEURIPS2021_252a3dba}, where $S^d_{ij} = d(V^i,V^j)$.  From layers 4 to 32, the representational similarity between neighbouring layers remains high (with scores above 0.95), suggesting that graph representational geometry changes relatively smoothly from early to late layers and does not appear to change in a clear step-like fashion as shown in Appendix Figure~\ref{fig:rsa_si}.

\textbf{Comparing LLM vs. hypothesis sequences.} To compare the model’s feedforward activations to classical search algorithms, we extract a possible sequence of algorithmic steps from the LLM hidden unit activations: for each layer $i$ we identified the node pair $e_i = (x,y)$ with highest representational similarity in $V_i$ (Figure. \ref{fig:caseSchema}). We compared this sequence of edges with all possible unique rollouts of BFS and DFS by: i) computing the edit distance between each LLM step and each BFS/DFS trajectory, and ii) finding the longest sequence of correct steps for each trajectory; to be maximally permissive, we only required LLM steps to be in the correct sequence, not necessarily in adjacent layers.  This analysis revealed that the sequence of steps identified in LLM layers is not well matched to any full trajectory under either BFS or DFS (the mean proportion of correct matching steps in sequence is $0.18$ for BFS and $0.24$ for DFS; Figure~\ref{fig:caseSchema}).

\textbf{Competing representations. } We next analyzed the evolution of node representations across layers. In Figure~\ref{fig:results}c, we present a low dimensional latent space (t-SNE) projection of node representations across all layers, along with the final end-of-sequence (`eos') representation (`yes' or `no').  
Colors represent room appearances (graph nodes) in the prompt, and numbers indicate the layer from which the representation is extracted, showing the evolution of representational distance among rooms over layers.

While in the first layers all nodes are closely clustered together, across the layers, we observe a clear progressive separation of tokens associated with the `lobby' from other nodes (Figure~\ref{fig:results}c). This  hints at an algorithmic strategy that distinguishes between the anchor tokens (the `lobby') and a set of varying potential goal nodes. Interestingly, we find further structure developing across the latter graph nodes: a consistent clustering of non-goal nodes, and a grouping of subsets among them as shown in Appendix Figure~\ref{fig:tsne2x2}.
Similar to our attention analysis results (Figure~\ref{fig:results}a), we observe that the competition between the goal node W and Q, W's closest competitor node in the same branch, is reflected in their progressively increasing separation in representation space (see Figure~\ref{fig:results}c and Appendix Figure~\ref{fig:tsne2x2}).
This separation typically emerges at intermediate layers and follows the high attention that W receives from the final `eos' token at these stages (Figure~\ref{fig:results}b). Lastly, we observe that the representation of the goal token, defined in the prompt context, consistently maps onto a distinct trajectory, separating its representation from all other nodes while being closest to the `lobby' representation (purple goal branch in Figure~\ref{fig:results}c).

\textbf{Search strategies in larger models.} To test whether increased scale results in a different, potentially more robust algorithmic search strategy, we repeated our experiments on a ten times larger model (Llama-3.1-70B-Instruct). As shown in Figure~\ref{fig:tsne_70B} and Figure~\ref{fig:llama70b_attention_combined} in the Appendix, our results are consistent to those of the smaller model,  displaying similar patterns in both representation space and attention scores, but with a more pronounced separation between correct and incorrect trajectories. We further compared the sequence of states with the highest representation activation across layers to potential search sequences generated by BFS and DFS strategies, and found no differences in search strategy between the models as shown in Figure~\ref{fig:bfs_dfs_70B} in the Appendix. Neither sequence of representations closely matched the BFS or DFS rollouts on the target graph.

\textbf{Interpretation.}
Using the framework of AlgEval, we found that attentional and representational patterns in LLMs do not align neatly with classical search algorithms, highlighting the role of algorithmic explanations in  making such discrepancies interpretable. The attention patterns we observed (Figure~\ref{fig:results}b and Figure~\ref{fig:llama70b_attention_combined} in the Appendix) suggest that current models do not construct a full world model or perform exhaustive search. We further find that representations of graph nodes evolve layer by layer, increasing the representational distance among key nodes, e.g., the distance between the closest competing node and the goal node (W vs.\ Q, Figure~\ref{fig:results} and Figure~\ref{fig:tsne_70B}). Future studies should verify whether this competition-driven separation across layers, also observed in simpler settings \cite{wang2023interpretability}, represents an algorithmic primitive, and  whether it is driven by inhibition or mover attention heads, function vectors \cite{todd2024function}, or other mechanisms. Furthermore, it remains to be seen to what extent  a model’s search strategy varies with task and model complexity, especially in relation to its failure modes when navigating more complex graphs.

\section{Alternative Views}

Our proposal has some overlap with mechanistic interpretability~\citep{olah2020zoom,olsson2022context}, but differs with key distinctions. First, mechanistic interpretability often emphasizes bottom-up perspectives, even advocating for hypothesis-free circuit analysis~\citep{olah2023interp}. Second, while it is good to remain open-minded about hypotheses, it has been argued that interpreting data with an ‘innocent eye’ is not possible ~\citep{gershman2021just}. Thus, we advocate for combining top-down algorithmic hypotheses with bottom-up evaluation. Third, while mechanistic interpretability focuses on low-level circuits, AlgEval targets the algorithmic level of explanation~\citep{marr1982vision}, integrating primitives and their composition, from circuits to higher-level computations. This perspective also connects to work on neural algorithmic reasoning~\citep{velivckovic2021neural}, which aims to integrate algorithmic structure into neural architectures by operating in high-dimensional latent spaces while performing computations aligned with a specific target algorithm. Fourth, we contrast our approach with “AI-assisted interpretability,” in which AI systems explain other AI systems \cite{choi2024scaling,li2024eliciting,olah2023interp}. Although it has produced intriguing results, like automated neuron descriptions~\citep{choi2024scaling}, we maintain it cannot replace hypothesis-driven research. Designing rigorous algorithmic tests likely requires reasoning skills beyond current models, so human involvement remains vital. Finally, AlgEval may offer an alternative to the dominant scaling paradigm~\citep{bitterlesson2019} by systematically understanding emergent algorithms and embedding them into architectures, rather than relying solely on training data and scale.

\section{Discussion and Future Directions}

In this position paper, we advocate for systematic research into algorithms learned and used by generative AI. We introduced AlgEval as a framework to investigate algorithmic primitives, their composition, and the impact of architecture, parameters, and optimization. AlgEval extends to inference-time compute and how it depends on training data and objectives, with implications for both empirical design and theoretical understanding.

\textbf{Theoretical research.} Recent theoretical work has addressed hierarchical language learning in Transformers \cite{allenzhu2024physicslanguagemodels1}, what formal languages they can learn \cite{Strobl_2024}, their representational strengths and limitations \cite{sanford2023representationalstrengthslimitationstransformers}, how they learn shortcuts to automata \cite{liu2023transformerslearnshortcutsautomata}, and how chain-of-thought provably improves their computation \cite{malach2023auto,li2024chain}. However, this work remains disjointed from interpretability research, and approaches to mechanistic understanding often lack formal theory. This makes connecting high-level explanations to low-level processes and generalizing insights across architectures difficult. Meanwhile, multi-agent systems show empirical benefits but remain theoretically under-explored, particularly regarding how agents interact, which algorithms they perform, and how their strategies differ from end-to-end models. Overall, theoretical work on the algorithmic capacities of LLMs is still sparse. Future studies should investigate properties of good solutions (axiomatic desiderata), optimality and complexity proofs, and learning theory for LLM algorithms. A key question is whether certain architectures, parameters, and optimization schemes can guarantee the implementation of specific algorithms. 

Algorithm evaluation should consider a number of relevant desiderata: \textit{Fidelity}, which ensures consistent input–output reliability, \textit{optimality}, which pursues the most efficient solutions \cite{10.1145/1011767.1011781}, and \textit{minimality}, which values lower algorithmic complexity \cite{elmoznino2024complexity}. Further, algorithms should be \textit{expressive} enough to be clearly understood and \textit{runtime efficient} for scalability \cite{10.1007/978-3-642-77927-5_24}. These properties form a basis for evaluating and developing algorithms in ML systems. An important direction is algorithm-centric architecture design and the reuse of verifiable algorithmic components, e.g., exploring symbolic operations to evoke specific algorithms. Another direction is to incentivize algorithmic building blocks during training, e.g., through complexity or sparsity regularization.

\textbf{Steering, guiding, optimization.} Optimizing a given LLM to learn or use specific algorithms can be steered via training, architecture, or ICL. Future work should examine how ICL can steer models toward a target algorithm and how feedforward or attention-based modifications improve primitives or composition. Another possibility is inference-time optimization to encourage specific algorithms. Ultimately, we can train or fine-tune LLMs to implement specific algorithms, guided by an algorithmic perspective.

\textbf{Designing new architectures.}
A key future direction is algorithm-centric architecture design, which may be accomplished through the identification and re-use of verifiable algorithmic components. It would be especially interesting to see, for instance, if specific algorithms can be evoked through the use of symbolic operations, in context learning, or training regimes, e.g., GRPO \cite{shao2024deepseekmathpushinglimitsmathematical} as opposed to Proximal Policy Optimization Algorithms or PPO \cite{schulman2017proximalpolicyoptimizationalgorithms}. An algorithmic understanding of architectural choices can improve how we build end-to-end architectures and multi-agent AI systems \cite{webb2024LLMPFC}, paving the way for more transparent, reliable, and theoretically grounded generative AI. This enhanced algorithmic understanding can have significant implications for scientific applications of ML, potentially uncovering efficient strategies and fundamental mechanisms across various domain sciences.

 \textbf{Conclusion.} While interpretability research has begun moving toward mechanistic and circuit-level analysis, it largely overlooks algorithmic explanations and evaluation of LLMs. \textbf{Our position is that the next generation of ML researchers should prioritize algorithmic understanding of generative AI.} We have presented AlgEval, approaches for algorithmic research, and actionable next steps. Deepening our understanding of how LLMs compute can enhance their sample efficiency, reduce emissions, and improve safety compliance, ultimately strengthening the overall impact of our community's work.





\section*{Acknowledgements}
The authors acknowledge Eder Sousa, Aishni Parab, Dalal Alharthi, Montassir Abbas, Andrea Kang, Hongjing Lu, Mark Green, Peter Todd, and the participants of the Institute for Pure and Applied Mathematics (IPAM) Long Program on the Mathematics of Intelligences.
Part of this research was performed while some of the authors were visiting IPAM, which is supported by the National Science Foundation (Grant No. DMS-1925919). 

\section*{Impact Statement}
We propose prioritizing a systematic understanding of algorithmic primitives and compositions in LLMs, and how they are learned and used given their architecture parameters, and training data. Below are a number of near-term and long-term impacts that can further motivate this research priority. We believe this research will impact more than mere understanding, and could potentially lead toward more efficient and lower-emission training and improving generative AI, and designing new architectures with algorithms in mind.

The rapid success and rise of generative AI faces the challenges of unpredictable errors as well as large emissions. In spite of these known challenges, the ML community's research priorities in the area of generative AI and LLMs have so far been myopically focused on improving performance, regardless of costs, and mechanistic interpretability, regardless of guarantees or algorithmic understanding. This position paper argues that the ML community should shift its focus to the algorithmic level of analysis, as the current priorities are often wasteful and offer only limited insight into models’ fundamental inner workings.

A previous position paper on inner interpretability \cite{pmlr-v235-vilas24a}, inspired by the analogy to Marr's levels in neuroscience \cite{marr1982vision}, calls for attention to all levels: the computational level, the algorithmic level, and the implementation level. While we agree, we think there is a specific need to prioritize an algorithmic understanding in line with long-standing traditions in computer science. We highlight the importance of putting resources into understanding the algorithmic vocabulary and grammar of generative AI, which will contribute to the field beyond understanding isolated phenomena and toward a more systematic foundation.

\subsection*{Sample Efficiency and Improved Behavior} 
An algorithmic understanding of generative AI can empower us to identify methods for improving sample-efficiency. This could occur through improving training with algorithmic performance in mind, optimizing and scaling inference-time compute and chain of thoughts with an understanding of their algorithmic implications. The latter could in turn improve the nebulous state of prompt engineering. Moreover, algorithm-based training and the reuse of algorithmic components in future models, both end-to-end and multi-agent, could lead to better behavioral performance and improved generalization. Ideally, this would yield fewer iterations for a given task, especially in multi-agent architectures.
Together, sample efficiency during training and compute efficiency during response generation could potentially address both compute and emission challenges of generative AI.
\subsection*{Emission Efficiency} Many assume generative AI’s emission costs occur only during training, which equate to the lifetime emissions of multiple cars \cite{strubell2019energypolicyconsiderationsdeep}. However, significant costs arise during use as well. For instance, a single interaction with a state-of-the-art LLM can require half a liter of water for cooling and emit the carbon equivalent of five gallons of gas when solving challenging problems with OpenAI’s GPT-3. It is estimated that by 2027, global AI’s water usage could rise to nearly two-thirds of the United Kingdom's annual consumption \cite{li2025makingaithirstyuncovering}. Repeated errors and back-and-forth interactions further increase these water and carbon costs, potentially leading to catastrophic climate impacts when scaled globally.
Adopting algorithm-driven approaches can enhance emission efficiency in training and reasoning, and inform the design of new architectures. We believe that an algorithmic understanding of LLMs can lead to more environmentally sustainable methods, which is crucial as these models are rapidly integrated into everyday products.

We believe an algorithmic understanding of LLMs can lead to less wasteful approaches, considering the environmental costs of training and using generative AI. This is especially important given the rapid pace at which these models are being integrated into everyday products. 

\subsection*{Theories for Multi-Agent AI} Given the unreliability of most generative AI approaches, multi-agent LLM architectures have become common to correct LLM errors and hallucinations, orchestrate actions, and improve reasoning, among other use cases. While these approaches make LLM-based solutions more reliable, they lead to even higher emissions. On the other hand, various approaches are a patch-work of popular knowledge of psychology, and at best cognitive science or brain-inspired approaches, without any systematic theories or guarantees. One of the key challenges lies in building and analyzing agentic systems with provable performance, ensuring that they can function reliably and effectively in complex environments. 
We believe a better understanding of what LLMs truly do can lead to more efficient design of multi-agent systems as well as their implementation and product integration, with advantages for engineers, users, and the planet.

\subsection*{Trust, Compliance, and Safety}
An enhanced understanding of model behavior enables researchers to discover novel mechanisms and advance a principled understanding of generative AI. An algorithmic understanding of LLMs can, in turn, increase the interoperability and trustworthiness of models. These insights can empower researchers and engineers to ensure compliance with safety standards.

\subsection*{Algorithmic Bias}
The challenge of bias in computer systems \cite{friedman1996}, particularly instances of \textit{technical bias} require an in-depth understanding of the underlying systems. In the context of generative AI, today's systems are commonly found to make unfair, undesired and even harmful predictions \cite{shah-etal-2020-predictive, lucy-bamman-2021-gender, contrastivebias2023, fang2024bias}, raising concerns about their deployment in sensitive domains. AlgEval directly supports the detection and understanding of algorithmic bias by systematically evaluating and interpreting a model’s fundamental components. As bias can manifest at different levels within a system, for example, in low-level primitives, compositions thereof, or in the functioning of full algorithms, AlgEval offers a complementary approach to addressing pre-existing bias originating from training data. By targeting technical bias at these distinct levels, it enables a more granular and thorough evaluation of bias in generative models.

\bibliography{main}

\begin{thebibliography}{140}
\providecommand{\natexlab}[1]{#1}
\providecommand{\url}[1]{\texttt{#1}}
\expandafter\ifx\csname urlstyle\endcsname\relax
  \providecommand{\doi}[1]{doi: #1}\else
  \providecommand{\doi}{doi: \begingroup \urlstyle{rm}\Url}\fi

\bibitem[Abnar \& Zuidema(2020)Abnar and Zuidema]{abnar-zuidema-2020-quantifying}
Abnar, S. and Zuidema, W.
\newblock Quantifying attention flow in transformers.
\newblock In Jurafsky, D., Chai, J., Schluter, N., and Tetreault, J. (eds.), \emph{Proceedings of the 58th Annual Meeting of the Association for Computational Linguistics}, pp.\  4190--4197, Online, July 2020. Association for Computational Linguistics.
\newblock \doi{10.18653/v1/2020.acl-main.385}.
\newblock URL \url{https://aclanthology.org/2020.acl-main.385/}.

\bibitem[Achtibat et~al.(2024)Achtibat, Hatefi, Dreyer, Jain, Wiegand, Lapuschkin, and Samek]{pmlr-v235-achtibat24a}
Achtibat, R., Hatefi, S. M.~V., Dreyer, M., Jain, A., Wiegand, T., Lapuschkin, S., and Samek, W.
\newblock {A}ttn{LRP}: Attention-aware layer-wise relevance propagation for transformers.
\newblock In Salakhutdinov, R., Kolter, Z., Heller, K., Weller, A., Oliver, N., Scarlett, J., and Berkenkamp, F. (eds.), \emph{Proceedings of the 41st International Conference on Machine Learning}, volume 235 of \emph{Proceedings of Machine Learning Research}, pp.\  135--168. PMLR, 21--27 Jul 2024.
\newblock URL \url{https://proceedings.mlr.press/v235/achtibat24a.html}.

\bibitem[Ali et~al.(2022)Ali, Schnake, Eberle, Montavon, M{\"{u}}ller, and Wolf]{transformerlrp2022}
Ali, A., Schnake, T., Eberle, O., Montavon, G., M{\"{u}}ller, K.-R., and Wolf, L.
\newblock {XAI} for transformers: Better explanations through conservative propagation.
\newblock In \emph{International Conference on Machine Learning, {ICML} 2022, 17-23 July 2022, Baltimore, Maryland, {USA}}, volume 162 of \emph{Proceedings of Machine Learning Research}, pp.\  435--451. {PMLR}, 2022.
\newblock URL \url{https://proceedings.mlr.press/v162/ali22a.html}.

\bibitem[Allen-Zhu \& Li(2024)Allen-Zhu and Li]{allenzhu2024physicslanguagemodels1}
Allen-Zhu, Z. and Li, Y.
\newblock Physics of language models: Part 1, learning hierarchical language structures, 2024.
\newblock URL \url{https://arxiv.org/abs/2305.13673}.

\bibitem[Baehrens et~al.(2010)Baehrens, Schroeter, Harmeling, Kawanabe, Hansen, and M{\"u}ller]{baehrens2010explain}
Baehrens, D., Schroeter, T., Harmeling, S., Kawanabe, M., Hansen, K., and M{\"u}ller, K.-R.
\newblock How to explain individual classification decisions.
\newblock \emph{The Journal of Machine Learning Research}, 11:\penalty0 1803--1831, 2010.

\bibitem[Besta et~al.(2024)Besta, Scheidl, Gianinazzi, Kwasniewski, Klaiman, Müller, and Hoefler]{besta2024demystifyinghigherordergraphneural}
Besta, M., Scheidl, F., Gianinazzi, L., Kwasniewski, G., Klaiman, S., Müller, J., and Hoefler, T.
\newblock Demystifying higher-order graph neural networks, 2024.
\newblock URL \url{https://arxiv.org/abs/2406.12841}.

\bibitem[Brunec \& Momennejad(2021)Brunec and Momennejad]{Brunec2021Predictive}
Brunec, I. and Momennejad, I.
\newblock Predictive representations in hippocampal and prefrontal hierarchies.
\newblock \emph{Journal of Neuroscience}, November 2021.
\newblock URL \url{https://www.jneurosci.org/content/early/2021/11/19/JN-RM-1327-21}.
\newblock JN-RM-1327-21.

\bibitem[Bürger et~al.(2024)Bürger, Hamprecht, and Nadler]{bürger2024truthuniversalrobustdetection}
Bürger, L., Hamprecht, F.~A., and Nadler, B.
\newblock Truth is universal: Robust detection of lies in llms, 2024.
\newblock URL \url{https://arxiv.org/abs/2407.12831}.

\bibitem[Campbell et~al.(2024)Campbell, Rane, Giallanza, De~Sabbata, Ghods, Joshi, Ku, Frankland, Griffiths, Cohen, et~al.]{campbell2024understanding}
Campbell, D., Rane, S., Giallanza, T., De~Sabbata, N., Ghods, K., Joshi, A., Ku, A., Frankland, S.~M., Griffiths, T.~L., Cohen, J.~D., et~al.
\newblock Understanding the limits of vision language models through the lens of the binding problem.
\newblock \emph{arXiv preprint arXiv:2411.00238}, 2024.

\bibitem[Cern{\'y}(1985)]{cerny1985thermodynamical}
Cern{\'y}, V.
\newblock Thermodynamical approach to the traveling salesman problem: An efficient simulation algorithm.
\newblock \emph{Journal of Optimization Theory and Applications}, 45\penalty0 (1):\penalty0 41--51, 1985.
\newblock \doi{10.1007/BF00940812}.

\bibitem[Chefer et~al.(2021)Chefer, Gur, and Wolf]{9577970}
Chefer, H., Gur, S., and Wolf, L.
\newblock Transformer interpretability beyond attention visualization.
\newblock In \emph{2021 IEEE/CVF Conference on Computer Vision and Pattern Recognition (CVPR)}, pp.\  782--791, 2021.
\newblock \doi{10.1109/CVPR46437.2021.00084}.

\bibitem[Choi et~al.(2024)Choi, Huang, Meng, Johnson, Steinhardt, and Schwettmann]{choi2024scaling}
Choi, D., Huang, V., Meng, K., Johnson, D., Steinhardt, J., and Schwettmann, S.
\newblock {Scaling Automatic Neuron Description}, 2024.
\newblock URL \url{https://transluce.org/neuron-descriptions}.

\bibitem[Chormai et~al.(2024)Chormai, Herrmann, M{\"u}ller, and Montavon]{chormai2024disentangled}
Chormai, P., Herrmann, J., M{\"u}ller, K.-R., and Montavon, G.
\newblock Disentangled explanations of neural network predictions by finding relevant subspaces.
\newblock \emph{IEEE Transactions on Pattern Analysis and Machine Intelligence}, 2024.

\bibitem[Clark et~al.(2019)Clark, Khandelwal, Levy, and Manning]{clark-etal-2019-bert}
Clark, K., Khandelwal, U., Levy, O., and Manning, C.~D.
\newblock What does {BERT} look at? an analysis of {BERT}`s attention.
\newblock In Linzen, T., Chrupa{\l}a, G., Belinkov, Y., and Hupkes, D. (eds.), \emph{Proceedings of the 2019 ACL Workshop BlackboxNLP: Analyzing and Interpreting Neural Networks for NLP}, pp.\  276--286, Florence, Italy, August 2019. Association for Computational Linguistics.
\newblock \doi{10.18653/v1/W19-4828}.
\newblock URL \url{https://aclanthology.org/W19-4828/}.

\bibitem[Conmy et~al.(2023)Conmy, Mavor-Parker, Lynch, Heimersheim, and Garriga-Alonso]{acdc2024}
Conmy, A., Mavor-Parker, A., Lynch, A., Heimersheim, S., and Garriga-Alonso, A.
\newblock Towards automated circuit discovery for mechanistic interpretability.
\newblock In Oh, A., Naumann, T., Globerson, A., Saenko, K., Hardt, M., and Levine, S. (eds.), \emph{Advances in Neural Information Processing Systems}, volume~36, pp.\  16318--16352. Curran Associates, Inc., 2023.

\bibitem[Conneau et~al.(2018)Conneau, Kruszewski, Lample, Barrault, and Baroni]{conneau-etal-2018-cram}
Conneau, A., Kruszewski, G., Lample, G., Barrault, L., and Baroni, M.
\newblock What you can cram into a single {\$}{\&}!{\#}* vector: Probing sentence embeddings for linguistic properties.
\newblock In Gurevych, I. and Miyao, Y. (eds.), \emph{Proceedings of the 56th Annual Meeting of the Association for Computational Linguistics (Volume 1: Long Papers)}, pp.\  2126--2136, Melbourne, Australia, July 2018. Association for Computational Linguistics.
\newblock \doi{10.18653/v1/P18-1198}.
\newblock URL \url{https://aclanthology.org/P18-1198/}.

\bibitem[Conwell et~al.(2024)Conwell, Tawiah-Quashie, and Ullman]{conwell2024relations}
Conwell, C., Tawiah-Quashie, R., and Ullman, T.
\newblock Relations, negations, and numbers: Looking for logic in generative text-to-image models.
\newblock \emph{arXiv preprint arXiv:2411.17066}, 2024.

\bibitem[DeepSeek-AI et~al.(2025)DeepSeek-AI, Guo, Yang, Zhang, Song, Zhang, Xu, Zhu, Ma, Wang, Bi, Zhang, Yu, Wu, Wu, Gou, Shao, Li, Gao, Liu, Xue, Wang, Wu, Feng, Lu, Zhao, Deng, Zhang, Ruan, Dai, Chen, Ji, Li, Lin, Dai, Luo, Hao, Chen, Li, Zhang, Bao, Xu, Wang, Ding, Xin, Gao, Qu, Li, Guo, Li, Wang, Chen, Yuan, Qiu, Li, Cai, Ni, Liang, Chen, Dong, Hu, Gao, Guan, Huang, Yu, Wang, Zhang, Zhao, Wang, Zhang, Xu, Xia, Zhang, Zhang, Tang, Li, Wang, Li, Tian, Huang, Zhang, Wang, Chen, Du, Ge, Zhang, Pan, Wang, Chen, Jin, Chen, Lu, Zhou, Chen, Ye, Wang, Yu, Zhou, Pan, Li, Zhou, Wu, Ye, Yun, Pei, Sun, Wang, Zeng, Zhao, Liu, Liang, Gao, Yu, Zhang, Xiao, An, Liu, Wang, Chen, Nie, Cheng, Liu, Xie, Liu, Yang, Li, Su, Lin, Li, Jin, Shen, Chen, Sun, Wang, Song, Zhou, Wang, Shan, Li, Wang, Wei, Zhang, Xu, Li, Zhao, Sun, Wang, Yu, Zhang, Shi, Xiong, He, Piao, Wang, Tan, Ma, Liu, Guo, Ou, Wang, Gong, Zou, He, Xiong, Luo, You, Liu, Zhou, Zhu, Xu, Huang, Li, Zheng, Zhu, Ma, Tang, Zha, Yan, Ren, Ren, Sha, Fu, Xu, Xie, Zhang,
  Hao, Ma, Yan, Wu, Gu, Zhu, Liu, Li, Xie, Song, Pan, Huang, Xu, Zhang, and Zhang]{deepseekai2025deepseekr1incentivizingreasoningcapability}
DeepSeek-AI, Guo, D., Yang, D., Zhang, H., Song, J., Zhang, R., Xu, R., Zhu, Q., Ma, S., Wang, P., Bi, X., Zhang, X., Yu, X., Wu, Y., Wu, Z.~F., Gou, Z., Shao, Z., Li, Z., Gao, Z., Liu, A., Xue, B., Wang, B., Wu, B., Feng, B., Lu, C., Zhao, C., Deng, C., Zhang, C., Ruan, C., Dai, D., Chen, D., Ji, D., Li, E., Lin, F., Dai, F., Luo, F., Hao, G., Chen, G., Li, G., Zhang, H., Bao, H., Xu, H., Wang, H., Ding, H., Xin, H., Gao, H., Qu, H., Li, H., Guo, J., Li, J., Wang, J., Chen, J., Yuan, J., Qiu, J., Li, J., Cai, J.~L., Ni, J., Liang, J., Chen, J., Dong, K., Hu, K., Gao, K., Guan, K., Huang, K., Yu, K., Wang, L., Zhang, L., Zhao, L., Wang, L., Zhang, L., Xu, L., Xia, L., Zhang, M., Zhang, M., Tang, M., Li, M., Wang, M., Li, M., Tian, N., Huang, P., Zhang, P., Wang, Q., Chen, Q., Du, Q., Ge, R., Zhang, R., Pan, R., Wang, R., Chen, R.~J., Jin, R.~L., Chen, R., Lu, S., Zhou, S., Chen, S., Ye, S., Wang, S., Yu, S., Zhou, S., Pan, S., Li, S.~S., Zhou, S., Wu, S., Ye, S., Yun, T., Pei, T., Sun, T., Wang, T., Zeng, W.,
  Zhao, W., Liu, W., Liang, W., Gao, W., Yu, W., Zhang, W., Xiao, W.~L., An, W., Liu, X., Wang, X., Chen, X., Nie, X., Cheng, X., Liu, X., Xie, X., Liu, X., Yang, X., Li, X., Su, X., Lin, X., Li, X.~Q., Jin, X., Shen, X., Chen, X., Sun, X., Wang, X., Song, X., Zhou, X., Wang, X., Shan, X., Li, Y.~K., Wang, Y.~Q., Wei, Y.~X., Zhang, Y., Xu, Y., Li, Y., Zhao, Y., Sun, Y., Wang, Y., Yu, Y., Zhang, Y., Shi, Y., Xiong, Y., He, Y., Piao, Y., Wang, Y., Tan, Y., Ma, Y., Liu, Y., Guo, Y., Ou, Y., Wang, Y., Gong, Y., Zou, Y., He, Y., Xiong, Y., Luo, Y., You, Y., Liu, Y., Zhou, Y., Zhu, Y.~X., Xu, Y., Huang, Y., Li, Y., Zheng, Y., Zhu, Y., Ma, Y., Tang, Y., Zha, Y., Yan, Y., Ren, Z.~Z., Ren, Z., Sha, Z., Fu, Z., Xu, Z., Xie, Z., Zhang, Z., Hao, Z., Ma, Z., Yan, Z., Wu, Z., Gu, Z., Zhu, Z., Liu, Z., Li, Z., Xie, Z., Song, Z., Pan, Z., Huang, Z., Xu, Z., Zhang, Z., and Zhang, Z.
\newblock Deepseek-r1: Incentivizing reasoning capability in llms via reinforcement learning, 2025.
\newblock URL \url{https://arxiv.org/abs/2501.12948}.

\bibitem[Demircan et~al.(2024)Demircan, Saanum, Jagadish, Binz, and Schulz]{demircan2024tdsaes}
Demircan, C., Saanum, T., Jagadish, A.~K., Binz, M., and Schulz, E.
\newblock Sparse autoencoders reveal temporal difference learning in large language models, 2024.
\newblock URL \url{https://arxiv.org/abs/2410.01280}.

\bibitem[Eberle et~al.(2022)Eberle, Büttner, Kräutli, Müller, Valleriani, and Montavon]{interactions2022}
Eberle, O., Büttner, J., Kräutli, F., Müller, K.-R., Valleriani, M., and Montavon, G.
\newblock Building and interpreting deep similarity models.
\newblock \emph{IEEE Transactions on Pattern Analysis and Machine Intelligence}, 44\penalty0 (3):\penalty0 1149--1161, 2022.
\newblock \doi{10.1109/TPAMI.2020.3020738}.

\bibitem[Eberle et~al.(2023)Eberle, Chalkidis, Cabello, and Brandl]{contrastivebias2023}
Eberle, O., Chalkidis, I., Cabello, L., and Brandl, S.
\newblock Rather a nurse than a physician - contrastive explanations under investigation.
\newblock In Bouamor, H., Pino, J., and Bali, K. (eds.), \emph{Proceedings of the 2023 Conference on Empirical Methods in Natural Language Processing}, pp.\  6907--6920, Singapore, December 2023. Association for Computational Linguistics.
\newblock \doi{10.18653/v1/2023.emnlp-main.427}.
\newblock URL \url{https://aclanthology.org/2023.emnlp-main.427/}.

\bibitem[Edelman et~al.(2024)Edelman, Tsilivis, Edelman, eran malach, and Goel]{edelman2024the}
Edelman, E., Tsilivis, N., Edelman, B.~L., eran malach, and Goel, S.
\newblock The evolution of statistical induction heads: In-context learning markov chains.
\newblock In \emph{The Thirty-eighth Annual Conference on Neural Information Processing Systems}, 2024.
\newblock URL \url{https://openreview.net/forum?id=qaRT6QTIqJ}.

\bibitem[Elhage et~al.(2021)Elhage, Nanda, Olsson, Henighan, Joseph, Mann, Askell, Bai, Chen, Conerly, DasSarma, Drain, Ganguli, Hatfield-Dodds, Hernandez, Jones, Kernion, Lovitt, Ndousse, Amodei, Brown, Clark, Kaplan, McCandlish, and Olah]{elhage2021mathematical}
Elhage, N., Nanda, N., Olsson, C., Henighan, T., Joseph, N., Mann, B., Askell, A., Bai, Y., Chen, A., Conerly, T., DasSarma, N., Drain, D., Ganguli, D., Hatfield-Dodds, Z., Hernandez, D., Jones, A., Kernion, J., Lovitt, L., Ndousse, K., Amodei, D., Brown, T., Clark, J., Kaplan, J., McCandlish, S., and Olah, C.
\newblock A mathematical framework for transformer circuits.
\newblock \emph{Transformer Circuits Thread}, 2021.
\newblock https://transformer-circuits.pub/2021/framework/index.html.

\bibitem[Elmoznino et~al.(2024)Elmoznino, Jiralerspong, Bengio, and Lajoie]{elmoznino2024complexity}
Elmoznino, E., Jiralerspong, T., Bengio, Y., and Lajoie, G.
\newblock A complexity-based theory of compositionality.
\newblock \emph{arXiv preprint arXiv:2410.14817}, 2024.

\bibitem[Fang et~al.(2024)Fang, Che, Mao, Zhang, Zhao, and Zhao]{fang2024bias}
Fang, X., Che, S., Mao, M., Zhang, H., Zhao, M., and Zhao, X.
\newblock Bias of ai-generated content: an examination of news produced by large language models.
\newblock \emph{Scientific Reports}, 14\penalty0 (1):\penalty0 5224, 2024.

\bibitem[Feng \& Steinhardt(2024)Feng and Steinhardt]{feng2024how}
Feng, J. and Steinhardt, J.
\newblock How do language models bind entities in context?
\newblock In \emph{The Twelfth International Conference on Learning Representations}, 2024.
\newblock URL \url{https://openreview.net/forum?id=zb3b6oKO77}.

\bibitem[Friedman \& Nissenbaum(1996)Friedman and Nissenbaum]{friedman1996}
Friedman, B. and Nissenbaum, H.
\newblock Bias in computer systems.
\newblock \emph{ACM Trans. Inf. Syst.}, 14\penalty0 (3):\penalty0 330–347, July 1996.
\newblock ISSN 1046-8188.
\newblock \doi{10.1145/230538.230561}.
\newblock URL \url{https://doi.org/10.1145/230538.230561}.

\bibitem[Fu et~al.(2024)Fu, CHEN, Jia, and Sharan]{fu2024transformers}
Fu, D., CHEN, T., Jia, R., and Sharan, V.
\newblock Transformers learn higher-order optimization methods for in-context learning: A study with linear models, 2024.
\newblock URL \url{https://openreview.net/forum?id=YKzGrt3m2g}.

\bibitem[Fumagalli et~al.(2023)Fumagalli, Muschalik, Kolpaczki, H{\"u}llermeier, and Hammer]{fumagalli2023shapiq}
Fumagalli, F., Muschalik, M., Kolpaczki, P., H{\"u}llermeier, E., and Hammer, B.~E.
\newblock {SHAP}-{IQ}: Unified approximation of any-order shapley interactions.
\newblock In \emph{Thirty-seventh Conference on Neural Information Processing Systems}, 2023.
\newblock URL \url{https://openreview.net/forum?id=IEMLNF4gK4}.

\bibitem[Galke et~al.(2024)Galke, Ram, and Raviv]{Galke2024}
Galke, L., Ram, Y., and Raviv, L.
\newblock {Deep neural networks and humans both benefit from compositional language structure}.
\newblock \emph{{Nature Communications}}, 15:\penalty0 10816, 2024.
\newblock ISSN 2041-1723.
\newblock \doi{10.1038/s41467-024-55158-1}.

\bibitem[Gandhi et~al.(2024)Gandhi, Lee, Grand, Liu, Cheng, Sharma, and Goodman]{gandhi2024stream}
Gandhi, K., Lee, D., Grand, G., Liu, M., Cheng, W., Sharma, A., and Goodman, N.~D.
\newblock Stream of search (sos): Learning to search in language.
\newblock \emph{arXiv preprint arXiv:2404.03683}, 2024.

\bibitem[Geiger et~al.(2022)Geiger, Wu, Lu, Rozner, Kreiss, Icard, Goodman, and Potts]{pmlr-v162-geiger22a}
Geiger, A., Wu, Z., Lu, H., Rozner, J., Kreiss, E., Icard, T., Goodman, N., and Potts, C.
\newblock Inducing causal structure for interpretable neural networks.
\newblock In Chaudhuri, K., Jegelka, S., Song, L., Szepesvari, C., Niu, G., and Sabato, S. (eds.), \emph{Proceedings of the 39th International Conference on Machine Learning}, volume 162 of \emph{Proceedings of Machine Learning Research}, pp.\  7324--7338. PMLR, 17--23 Jul 2022.
\newblock URL \url{https://proceedings.mlr.press/v162/geiger22a.html}.

\bibitem[Gershman(2019)]{gershman2019amortized}
Gershman, S.~J.
\newblock Amortized inference in learning and decision making.
\newblock \emph{Current Opinion in Behavioral Sciences}, 29:\penalty0 80--86, 2019.
\newblock \doi{10.1016/j.cobeha.2019.04.003}.

\bibitem[Gershman(2021)]{gershman2021just}
Gershman, S.~J.
\newblock Just looking: The innocent eye in neuroscience.
\newblock \emph{Neuron}, 109\penalty0 (14):\penalty0 2220--2223, 2021.

\bibitem[Geva et~al.(2021)Geva, Schuster, Berant, and Levy]{geva-etal-2021-transformer}
Geva, M., Schuster, R., Berant, J., and Levy, O.
\newblock Transformer feed-forward layers are key-value memories.
\newblock In Moens, M.-F., Huang, X., Specia, L., and Yih, S. W.-t. (eds.), \emph{Proceedings of the 2021 Conference on Empirical Methods in Natural Language Processing}, pp.\  5484--5495, Online and Punta Cana, Dominican Republic, November 2021. Association for Computational Linguistics.
\newblock \doi{10.18653/v1/2021.emnlp-main.446}.
\newblock URL \url{https://aclanthology.org/2021.emnlp-main.446/}.

\bibitem[Giaffar et~al.(2024)Giaffar, Rull\'{a}n~Bux\'{o}, and Aoi]{ENSD_pmlr-v238-giaffar24a}
Giaffar, H., Rull\'{a}n~Bux\'{o}, C., and Aoi, M.
\newblock The effective number of shared dimensions between paired datasets.
\newblock In \emph{Proceedings of The 27th International Conference on Artificial Intelligence and Statistics}, volume 238 of \emph{Proceedings of Machine Learning Research}, pp.\  4249--4257. PMLR, 2024.
\newblock URL \url{https://proceedings.mlr.press/v238/giaffar24a.html}.

\bibitem[Gurnee \& Tegmark(2024)Gurnee and Tegmark]{gurnee2024languagemodelsrepresentspace}
Gurnee, W. and Tegmark, M.
\newblock Language models represent space and time, 2024.
\newblock URL \url{https://arxiv.org/abs/2310.02207}.

\bibitem[Hanna et~al.(2024)Hanna, Pezzelle, and Belinkov]{hanna2024have}
Hanna, M., Pezzelle, S., and Belinkov, Y.
\newblock Have faith in faithfulness: Going beyond circuit overlap when finding model mechanisms.
\newblock In \emph{ICML 2024 Workshop on Mechanistic Interpretability}, 2024.
\newblock URL \url{https://openreview.net/forum?id=grXgesr5dT}.

\bibitem[Hewitt \& Manning(2019)Hewitt and Manning]{hewitt-manning-2019-structural}
Hewitt, J. and Manning, C.~D.
\newblock {A} structural probe for finding syntax in word representations.
\newblock In Burstein, J., Doran, C., and Solorio, T. (eds.), \emph{Proceedings of the 2019 Conference of the North {A}merican Chapter of the Association for Computational Linguistics: Human Language Technologies, Volume 1 (Long and Short Papers)}, pp.\  4129--4138, Minneapolis, Minnesota, June 2019. Association for Computational Linguistics.
\newblock \doi{10.18653/v1/N19-1419}.
\newblock URL \url{https://aclanthology.org/N19-1419/}.

\bibitem[Huh et~al.(2024)Huh, Cheung, Wang, and Isola]{pmlr-v235-huh24a}
Huh, M., Cheung, B., Wang, T., and Isola, P.
\newblock Position: The platonic representation hypothesis.
\newblock In Salakhutdinov, R., Kolter, Z., Heller, K., Weller, A., Oliver, N., Scarlett, J., and Berkenkamp, F. (eds.), \emph{Proceedings of the 41st International Conference on Machine Learning}, volume 235 of \emph{Proceedings of Machine Learning Research}, pp.\  20617--20642. PMLR, 21--27 Jul 2024.
\newblock URL \url{https://proceedings.mlr.press/v235/huh24a.html}.

\bibitem[Jaech et~al.(2024)Jaech, Kalai, Lerer, Richardson, El-Kishky, Low, Helyar, Madry, Beutel, Carney, et~al.]{jaech2024openai}
Jaech, A., Kalai, A., Lerer, A., Richardson, A., El-Kishky, A., Low, A., Helyar, A., Madry, A., Beutel, A., Carney, A., et~al.
\newblock Openai o1 system card.
\newblock \emph{arXiv preprint arXiv:2412.16720}, 2024.

\bibitem[Jafari et~al.(2024)Jafari, Montavon, Müller, and Eberle]{jafari2024mambalrp}
Jafari, F.~R., Montavon, G., Müller, K.-R., and Eberle, O.
\newblock Mambalrp: Explaining selective state space sequence models.
\newblock In \emph{The Thirty-eighth Annual Conference on Neural Information Processing Systems}, 2024.

\bibitem[Kandpal et~al.(2023)Kandpal, Deng, Roberts, Wallace, and Raffel]{kandpal2023large}
Kandpal, N., Deng, H., Roberts, A., Wallace, E., and Raffel, C.
\newblock Large language models struggle to learn long-tail knowledge.
\newblock In \emph{International Conference on Machine Learning}, pp.\  15696--15707. PMLR, 2023.

\bibitem[Kauffmann et~al.(2022)Kauffmann, Esders, Ruff, Montavon, Samek, and M{\"u}ller]{kauffmann2022clustering}
Kauffmann, J., Esders, M., Ruff, L., Montavon, G., Samek, W., and M{\"u}ller, K.-R.
\newblock From clustering to cluster explanations via neural networks.
\newblock \emph{IEEE Transactions on Neural Networks and Learning Systems}, 35\penalty0 (2):\penalty0 1926--1940, 2022.

\bibitem[Kauffmann et~al.(2024)Kauffmann, Dippel, Ruff, Samek, M{\"u}ller, and Montavon]{kauffmann2024clever}
Kauffmann, J., Dippel, J., Ruff, L., Samek, W., M{\"u}ller, K.-R., and Montavon, G.
\newblock The clever hans effect in unsupervised learning.
\newblock \emph{arXiv preprint arXiv:2408.08041}, 2024.

\bibitem[Kaur et~al.(2022)Kaur, Uslu, Rittichier, and Durresi]{10.1145/3491209}
Kaur, D., Uslu, S., Rittichier, K.~J., and Durresi, A.
\newblock Trustworthy artificial intelligence: A review.
\newblock \emph{ACM Comput. Surv.}, 55\penalty0 (2), January 2022.
\newblock ISSN 0360-0300.
\newblock \doi{10.1145/3491209}.
\newblock URL \url{https://doi.org/10.1145/3491209}.

\bibitem[Kirkpatrick et~al.(1983)Kirkpatrick, Gelatt, and Vecchi]{kirkpatrick1983optimization}
Kirkpatrick, S., Gelatt, C.~D., and Vecchi, M.~P.
\newblock Optimization by simulated annealing.
\newblock \emph{Science}, 220\penalty0 (4598):\penalty0 671--680, 1983.
\newblock \doi{10.1126/science.220.4598.671}.

\bibitem[Klabunde et~al.(2024)Klabunde, Schumacher, Strohmaier, and Lemmerich]{klabunde2024similarityneuralnetworkmodels}
Klabunde, M., Schumacher, T., Strohmaier, M., and Lemmerich, F.
\newblock Similarity of neural network models: A survey of functional and representational measures, 2024.
\newblock URL \url{https://arxiv.org/abs/2305.06329}.

\bibitem[Knuth({1968})]{Knuth-1968}
Knuth, D.~E.
\newblock {Semantics of Context-Free Languages}.
\newblock \emph{{Mathematical Systems Theory}}, {2}\penalty0 ({2}):\penalty0 127--145, {1968}.

\bibitem[Kornblith et~al.(2019)Kornblith, Norouzi, Lee, and Hinton]{kornblith2019similarity}
Kornblith, S., Norouzi, M., Lee, H., and Hinton, G.
\newblock Similarity of neural network representations revisited.
\newblock In \emph{International conference on machine learning}, pp.\  3519--3529. PMLR, 2019.

\bibitem[Kriegeskorte et~al.(2008)Kriegeskorte, Mur, and Bandettini]{kriegeskorte2008representational}
Kriegeskorte, N., Mur, M., and Bandettini, P.~A.
\newblock Representational similarity analysis-connecting the branches of systems neuroscience.
\newblock \emph{Frontiers in systems neuroscience}, 2:\penalty0 249, 2008.

\bibitem[Lehnert et~al.(2024)Lehnert, Sukhbaatar, Su, Zheng, McVay, Rabbat, and Tian]{lehnert2024beyond}
Lehnert, L., Sukhbaatar, S., Su, D., Zheng, Q., McVay, P., Rabbat, M., and Tian, Y.
\newblock Beyond a*: Better planning with transformers via search dynamics bootstrapping.
\newblock In \emph{First Conference on Language Modeling}, 2024.
\newblock URL \url{https://openreview.net/forum?id=SGoVIC0u0f}.

\bibitem[Lepori et~al.(2023)Lepori, Serre, and Pavlick]{lepori2023break}
Lepori, M., Serre, T., and Pavlick, E.
\newblock Break it down: Evidence for structural compositionality in neural networks.
\newblock \emph{Advances in Neural Information Processing Systems}, 36:\penalty0 42623--42660, 2023.

\bibitem[Lewis et~al.(2022)Lewis, Nayak, Yu, Yu, Merullo, Bach, and Pavlick]{lewis2022does}
Lewis, M., Nayak, N.~V., Yu, P., Yu, Q., Merullo, J., Bach, S.~H., and Pavlick, E.
\newblock Does clip bind concepts? probing compositionality in large image models.
\newblock \emph{arXiv preprint arXiv:2212.10537}, 2022.

\bibitem[Li et~al.(2025)Li, Yang, Islam, and Ren]{li2025makingaithirstyuncovering}
Li, P., Yang, J., Islam, M.~A., and Ren, S.
\newblock Making ai less "thirsty": Uncovering and addressing the secret water footprint of ai models, 2025.
\newblock URL \url{https://arxiv.org/abs/2304.03271}.

\bibitem[Li et~al.(2024{\natexlab{a}})Li, Chowdhury, Johnson, Hashimoto, Liang, Schwettmann, and Steinhardt]{li2024eliciting}
Li, X., Chowdhury, N., Johnson, D., Hashimoto, T., Liang, P., Schwettmann, S., and Steinhardt, J.
\newblock {Eliciting Language Model Behaviors with Investigator Agents}, 2024{\natexlab{a}}.
\newblock URL \url{https://transluce.org/automated-elicitation}.

\bibitem[Li et~al.(2023)Li, Ildiz, Papailiopoulos, and Oymak]{li2023transformersalgorithmsgeneralizationstability}
Li, Y., Ildiz, M.~E., Papailiopoulos, D., and Oymak, S.
\newblock Transformers as algorithms: Generalization and stability in in-context learning, 2023.
\newblock URL \url{https://arxiv.org/abs/2301.07067}.

\bibitem[Li et~al.(2024{\natexlab{b}})Li, Liu, Zhou, and Ma]{li2024chain}
Li, Z., Liu, H., Zhou, D., and Ma, T.
\newblock Chain of thought empowers transformers to solve inherently serial problems.
\newblock \emph{arXiv preprint arXiv:2402.12875}, 2024{\natexlab{b}}.

\bibitem[Lin et~al.(2017)Lin, Dollár, Girshick, He, Hariharan, and Belongie]{pyramids2017}
Lin, T.-Y., Dollár, P., Girshick, R., He, K., Hariharan, B., and Belongie, S.
\newblock Feature pyramid networks for object detection.
\newblock In \emph{2017 IEEE Conference on Computer Vision and Pattern Recognition (CVPR)}, pp.\  936--944, 2017.
\newblock \doi{10.1109/CVPR.2017.106}.

\bibitem[Lipton(2017)]{lipton2017mythosmodelinterpretability}
Lipton, Z.~C.
\newblock The mythos of model interpretability, 2017.
\newblock URL \url{https://arxiv.org/abs/1606.03490}.

\bibitem[Liu et~al.(2023)Liu, Ash, Goel, Krishnamurthy, and Zhang]{liu2023transformerslearnshortcutsautomata}
Liu, B., Ash, J.~T., Goel, S., Krishnamurthy, A., and Zhang, C.
\newblock Transformers learn shortcuts to automata, 2023.
\newblock URL \url{https://arxiv.org/abs/2210.10749}.

\bibitem[Lucy \& Bamman(2021)Lucy and Bamman]{lucy-bamman-2021-gender}
Lucy, L. and Bamman, D.
\newblock Gender and representation bias in {GPT}-3 generated stories.
\newblock In Akoury, N., Brahman, F., Chaturvedi, S., Clark, E., Iyyer, M., and Martin, L.~J. (eds.), \emph{Proceedings of the Third Workshop on Narrative Understanding}, pp.\  48--55, Virtual, June 2021. Association for Computational Linguistics.
\newblock \doi{10.18653/v1/2021.nuse-1.5}.
\newblock URL \url{https://aclanthology.org/2021.nuse-1.5/}.

\bibitem[Lundberg \& Lee(2017)Lundberg and Lee]{NIPS2017_8a20a862}
Lundberg, S.~M. and Lee, S.-I.
\newblock A unified approach to interpreting model predictions.
\newblock In Guyon, I., Luxburg, U.~V., Bengio, S., Wallach, H., Fergus, R., Vishwanathan, S., and Garnett, R. (eds.), \emph{Advances in Neural Information Processing Systems}, volume~30. Curran Associates, Inc., 2017.
\newblock URL \url{https://proceedings.neurips.cc/paper_files/paper/2017/file/8a20a8621978632d76c43dfd28b67767-Paper.pdf}.

\bibitem[Malach(2023)]{malach2023auto}
Malach, E.
\newblock Auto-regressive next-token predictors are universal learners.
\newblock \emph{arXiv preprint arXiv:2309.06979}, 2023.

\bibitem[Manvi et~al.(2024)Manvi, Singh, and Ermon]{manvi2024adaptiveinferencetimecomputellms}
Manvi, R., Singh, A., and Ermon, S.
\newblock Adaptive inference-time compute: Llms can predict if they can do better, even mid-generation, 2024.
\newblock URL \url{https://arxiv.org/abs/2410.02725}.

\bibitem[Marr(1982)]{marr1982vision}
Marr, D.
\newblock \emph{Vision: A computational investigation into the human representation and processing of visual information}.
\newblock MIT press, 1982.

\bibitem[McDougall et~al.(2024)McDougall, Conmy, Rushing, McGrath, and Nanda]{mcdougall-etal-2024-copy}
McDougall, C.~S., Conmy, A., Rushing, C., McGrath, T., and Nanda, N.
\newblock Copy suppression: Comprehensively understanding a motif in language model attention heads.
\newblock In Belinkov, Y., Kim, N., Jumelet, J., Mohebbi, H., Mueller, A., and Chen, H. (eds.), \emph{Proceedings of the 7th BlackboxNLP Workshop: Analyzing and Interpreting Neural Networks for NLP}, pp.\  337--363, Miami, Florida, US, November 2024. Association for Computational Linguistics.
\newblock \doi{10.18653/v1/2024.blackboxnlp-1.22}.
\newblock URL \url{https://aclanthology.org/2024.blackboxnlp-1.22/}.

\bibitem[Merullo et~al.(2024)Merullo, Eickhoff, and Pavlick]{merullo-etal-2024-language}
Merullo, J., Eickhoff, C., and Pavlick, E.
\newblock Language models implement simple {W}ord2{V}ec-style vector arithmetic.
\newblock In Duh, K., Gomez, H., and Bethard, S. (eds.), \emph{Proceedings of the 2024 Conference of the North American Chapter of the Association for Computational Linguistics: Human Language Technologies (Volume 1: Long Papers)}, pp.\  5030--5047, Mexico City, Mexico, June 2024. Association for Computational Linguistics.
\newblock \doi{10.18653/v1/2024.naacl-long.281}.
\newblock URL \url{https://aclanthology.org/2024.naacl-long.281/}.

\bibitem[Metropolis et~al.(1953)Metropolis, Rosenbluth, Rosenbluth, Teller, and Teller]{metropolis1953equation}
Metropolis, N., Rosenbluth, A.~W., Rosenbluth, M.~N., Teller, A.~H., and Teller, E.
\newblock Equation of state calculations by fast computing machines.
\newblock \emph{The Journal of Chemical Physics}, 21\penalty0 (6):\penalty0 1087--1092, 1953.
\newblock \doi{10.1063/1.1699114}.

\bibitem[Mitchell et~al.(2023)Mitchell, Palmarini, and Moskvichev]{mitchell2023comparing}
Mitchell, M., Palmarini, A.~B., and Moskvichev, A.
\newblock Comparing humans, gpt-4, and gpt-4v on abstraction and reasoning tasks.
\newblock \emph{arXiv preprint arXiv:2311.09247}, 2023.

\bibitem[Momennejad(2020)]{Momennejad2020LearningStructures}
Momennejad, I.
\newblock Learning structures: Predictive representations, replay, and generalization.
\newblock \emph{Current Opinion in Behavioral Sciences}, 32:\penalty0 155--166, April 2020.
\newblock URL \url{https://www.sciencedirect.com/science/article/pii/S2352154620300371}.

\bibitem[Momennejad(forthcoming)]{momennejad_forthcoming}
Momennejad, I.
\newblock Memory and planning in brains and machines: Multiscale predictive representations.
\newblock In Nadel, L. and Aronovitz, S. (eds.), \emph{Space, Time, and Memory}. Oxford University Press, forthcoming.

\bibitem[Momennejad et~al.(2017)Momennejad, Russek, Cheong, Botvinick, Daw, and Gershman]{Momennejad2017Successor}
Momennejad, I., Russek, E., Cheong, J.~H., Botvinick, M.~M., Daw, N., and Gershman, S.~J.
\newblock The successor representation in human reinforcement learning: evidence from retrospective revaluation.
\newblock \emph{Nature Human Behaviour}, 1:\penalty0 680–692, 2017.
\newblock \doi{10.1038/s41562-017-0071}.
\newblock URL \url{https://www.nature.com/articles/s41562-017-0071}.
\newblock Equal contribution.

\bibitem[Momennejad et~al.(2023)Momennejad, Hasanbeig, Vieira, Sharma, Ness, Jojic, Palangi, and Larson]{momennejad2023cogeval}
Momennejad, I., Hasanbeig, H., Vieira, F., Sharma, H., Ness, R.~O., Jojic, N., Palangi, H., and Larson, J.
\newblock Evaluating cognitive maps and planning in large language models with cogeval, 2023.
\newblock URL \url{https://arxiv.org/abs/2309.15129}.

\bibitem[Montavon et~al.(2018)Montavon, Samek, and Müller]{MONTAVON20181}
Montavon, G., Samek, W., and Müller, K.-R.
\newblock Methods for interpreting and understanding deep neural networks.
\newblock \emph{Digital Signal Processing}, 73:\penalty0 1--15, 2018.
\newblock ISSN 1051-2004.
\newblock \doi{https://doi.org/10.1016/j.dsp.2017.10.011}.
\newblock URL \url{https://www.sciencedirect.com/science/article/pii/S1051200417302385}.

\bibitem[Morris et~al.(2019)Morris, Ritzert, Fey, Hamilton, Lenssen, Rattan, and Grohe]{morris2019weisfeiler}
Morris, C., Ritzert, M., Fey, M., Hamilton, W.~L., Lenssen, J.~E., Rattan, G., and Grohe, M.
\newblock Weisfeiler and leman go neural: higher-order graph neural networks.
\newblock In \emph{Proceedings of the Thirty-Third AAAI Conference on Artificial Intelligence and Thirty-First Innovative Applications of Artificial Intelligence Conference and Ninth AAAI Symposium on Educational Advances in Artificial Intelligence}, AAAI'19/IAAI'19/EAAI'19, pp.\  4602--4609. AAAI Press, 2019.
\newblock ISBN 978-1-57735-809-1.
\newblock \doi{10.1609/aaai.v33i01.33014602}.
\newblock URL \url{https://doi.org/10.1609/aaai.v33i01.33014602}.

\bibitem[Nanda et~al.(2023)Nanda, Chan, Lieberum, Smith, and Steinhardt]{nanda2023progress}
Nanda, N., Chan, L., Lieberum, T., Smith, J., and Steinhardt, J.
\newblock Progress measures for grokking via mechanistic interpretability.
\newblock In \emph{The Eleventh International Conference on Learning Representations}, 2023.
\newblock URL \url{https://openreview.net/forum?id=9XFSbDPmdW}.

\bibitem[Nisioti et~al.(2024)Nisioti, Risi, Momennejad, Oudeyer, and Moulin-Frier]{nisioti2024}
Nisioti, E., Risi, S., Momennejad, I., Oudeyer, P.-Y., and Moulin-Frier, C.
\newblock Collective innovation in groups of large language models, 2024.
\newblock URL \url{https://arxiv.org/abs/2407.05377}.

\bibitem[Olah(2023)]{olah2023interp}
Olah, C.
\newblock {Interpretability Dreams}, 2023.
\newblock URL \url{https://transformer-circuits.pub/2023/interpretability-dreams}.

\bibitem[Olah et~al.(2020)Olah, Cammarata, Schubert, Goh, Petrov, and Carter]{olah2020zoom}
Olah, C., Cammarata, N., Schubert, L., Goh, G., Petrov, M., and Carter, S.
\newblock Zoom in: An introduction to circuits.
\newblock \emph{Distill}, 2020.
\newblock \doi{10.23915/distill.00024.001}.
\newblock https://distill.pub/2020/circuits/zoom-in.

\bibitem[Olsson et~al.(2022)Olsson, Elhage, Nanda, Joseph, DasSarma, Henighan, Mann, Askell, Bai, Chen, Conerly, Drain, Ganguli, Hatfield-Dodds, Hernandez, Johnston, Jones, Kernion, Lovitt, Ndousse, Amodei, Brown, Clark, Kaplan, McCandlish, and Olah]{olsson2022context}
Olsson, C., Elhage, N., Nanda, N., Joseph, N., DasSarma, N., Henighan, T., Mann, B., Askell, A., Bai, Y., Chen, A., Conerly, T., Drain, D., Ganguli, D., Hatfield-Dodds, Z., Hernandez, D., Johnston, S., Jones, A., Kernion, J., Lovitt, L., Ndousse, K., Amodei, D., Brown, T., Clark, J., Kaplan, J., McCandlish, S., and Olah, C.
\newblock In-context learning and induction heads.
\newblock \emph{Transformer Circuits Thread}, 2022.
\newblock https://transformer-circuits.pub/2022/in-context-learning-and-induction-heads/index.html.

\bibitem[Power et~al.(2022)Power, Burda, Edwards, Babuschkin, and Misra]{power2022grokking}
Power, A., Burda, Y., Edwards, H., Babuschkin, I., and Misra, V.
\newblock Grokking: Generalization beyond overfitting on small algorithmic datasets.
\newblock \emph{arXiv preprint arXiv:2201.02177}, 2022.

\bibitem[Qiu et~al.(2022)Qiu, Shaw, Pasupat, Shi, Herzig, Pitler, Sha, and Toutanova]{qiu-etal-2022-evaluating}
Qiu, L., Shaw, P., Pasupat, P., Shi, T., Herzig, J., Pitler, E., Sha, F., and Toutanova, K.
\newblock Evaluating the impact of model scale for compositional generalization in semantic parsing.
\newblock In \emph{Proc. of Conf. on Empirical Methods in Natural Language Processing ({EMNLP})}, pp.\  9157--9179. ACL, December 2022.
\newblock \doi{10.18653/v1/2022.emnlp-main.624}.

\bibitem[Radev et~al.(2020)Radev, Voss, Wieschen, and Bürkner]{radev2020amortizedbayesianinferencemodels}
Radev, S.~T., Voss, A., Wieschen, E.~M., and Bürkner, P.-C.
\newblock Amortized bayesian inference for models of cognition, 2020.
\newblock URL \url{https://arxiv.org/abs/2005.03899}.

\bibitem[Ren et~al.(2024)Ren, Guo, Yan, Liu, Zhang, Qiu, and Lin]{ren-etal-2024-identifying}
Ren, J., Guo, Q., Yan, H., Liu, D., Zhang, Q., Qiu, X., and Lin, D.
\newblock Identifying semantic induction heads to understand in-context learning.
\newblock In Ku, L.-W., Martins, A., and Srikumar, V. (eds.), \emph{Findings of the Association for Computational Linguistics: ACL 2024}, pp.\  6916--6932, Bangkok, Thailand, August 2024. Association for Computational Linguistics.
\newblock \doi{10.18653/v1/2024.findings-acl.412}.
\newblock URL \url{https://aclanthology.org/2024.findings-acl.412/}.

\bibitem[Russek et~al.(2017)Russek, Momennejad, Botvinick, Gershman, and Daw]{Russek2017Predictive}
Russek, E., Momennejad, I., Botvinick, M.~M., Gershman, S.~J., and Daw, N.
\newblock Predictive representations can link model-based reinforcement learning to model-free mechanisms.
\newblock \emph{PLOS Computational Biology}, 13\penalty0 (6):\penalty0 e1005474, 2017.
\newblock \doi{10.1371/journal.pcbi.1005474}.
\newblock URL \url{https://journals.plos.org/ploscompbiol/article?id=10.1371/journal.pcbi.1005474}.

\bibitem[Samek et~al.(2021)Samek, Montavon, Lapuschkin, Anders, and Müller]{samek2021}
Samek, W., Montavon, G., Lapuschkin, S., Anders, C.~J., and Müller, K.-R.
\newblock Explaining deep neural networks and beyond: A review of methods and applications.
\newblock \emph{Proceedings of the IEEE}, 109\penalty0 (3):\penalty0 247--278, 2021.
\newblock \doi{10.1109/JPROC.2021.3060483}.

\bibitem[Sanford et~al.(2023)Sanford, Hsu, and Telgarsky]{sanford2023representationalstrengthslimitationstransformers}
Sanford, C., Hsu, D., and Telgarsky, M.
\newblock Representational strengths and limitations of transformers, 2023.
\newblock URL \url{https://arxiv.org/abs/2306.02896}.

\bibitem[Sanford et~al.(2024)Sanford, Fatemi, Hall, Tsitsulin, Kazemi, Halcrow, Perozzi, and Mirrokni]{sanford2024understanding}
Sanford, C., Fatemi, B., Hall, E., Tsitsulin, A., Kazemi, M., Halcrow, J., Perozzi, B., and Mirrokni, V.
\newblock Understanding transformer reasoning capabilities via graph algorithms.
\newblock In \emph{The Thirty-eighth Annual Conference on Neural Information Processing Systems}, 2024.
\newblock URL \url{https://openreview.net/forum?id=AfzbDw6DSp}.

\bibitem[Satorras et~al.(2021)Satorras, Hoogeboom, and Welling]{satorras2021n}
Satorras, V.~G., Hoogeboom, E., and Welling, M.
\newblock E(n) equivariant graph neural networks.
\newblock In \emph{International conference on machine learning}, pp.\  9323--9332. PMLR, 2021.

\bibitem[Schnake et~al.(2022)Schnake, Eberle, Lederer, Nakajima, Schütt, Müller, and Montavon]{higherorder2022}
Schnake, T., Eberle, O., Lederer, J., Nakajima, S., Schütt, K.~T., Müller, K.-R., and Montavon, G.
\newblock Higher-order explanations of graph neural networks via relevant walks.
\newblock \emph{IEEE Transactions on Pattern Analysis and Machine Intelligence}, 44\penalty0 (11):\penalty0 7581--7596, 2022.
\newblock \doi{10.1109/TPAMI.2021.3115452}.

\bibitem[Schulman et~al.(2017)Schulman, Wolski, Dhariwal, Radford, and Klimov]{schulman2017proximalpolicyoptimizationalgorithms}
Schulman, J., Wolski, F., Dhariwal, P., Radford, A., and Klimov, O.
\newblock Proximal policy optimization algorithms, 2017.
\newblock URL \url{https://arxiv.org/abs/1707.06347}.

\bibitem[Shah et~al.(2020)Shah, Schwartz, and Hovy]{shah-etal-2020-predictive}
Shah, D.~S., Schwartz, H.~A., and Hovy, D.
\newblock Predictive biases in natural language processing models: A conceptual framework and overview.
\newblock In Jurafsky, D., Chai, J., Schluter, N., and Tetreault, J. (eds.), \emph{Proceedings of the 58th Annual Meeting of the Association for Computational Linguistics}, pp.\  5248--5264, Online, July 2020. Association for Computational Linguistics.
\newblock \doi{10.18653/v1/2020.acl-main.468}.
\newblock URL \url{https://aclanthology.org/2020.acl-main.468/}.

\bibitem[Shao et~al.(2024)Shao, Wang, Zhu, Xu, Song, Bi, Zhang, Zhang, Li, Wu, and Guo]{shao2024deepseekmathpushinglimitsmathematical}
Shao, Z., Wang, P., Zhu, Q., Xu, R., Song, J., Bi, X., Zhang, H., Zhang, M., Li, Y.~K., Wu, Y., and Guo, D.
\newblock Deepseekmath: Pushing the limits of mathematical reasoning in open language models, 2024.
\newblock URL \url{https://arxiv.org/abs/2402.03300}.

\bibitem[Sharma et~al.(2023)Sharma, Cz{\'e}gel, Lachmann, Kempes, Walker, and Cronin]{sharma2023assembly}
Sharma, A., Cz{\'e}gel, D., Lachmann, M., Kempes, C.~P., Walker, S.~I., and Cronin, L.
\newblock Assembly theory explains and quantifies selection and evolution.
\newblock \emph{Nature}, 622\penalty0 (7982):\penalty0 321--328, 2023.

\bibitem[Shneidman \& Parkes(2004)Shneidman and Parkes]{10.1145/1011767.1011781}
Shneidman, J. and Parkes, D.~C.
\newblock Specification faithfulness in networks with rational nodes.
\newblock In \emph{Proceedings of the Twenty-Third Annual ACM Symposium on Principles of Distributed Computing}, PODC '04, pp.\  88–97, New York, NY, USA, 2004. Association for Computing Machinery.
\newblock ISBN 1581138024.
\newblock \doi{10.1145/1011767.1011781}.
\newblock URL \url{https://doi.org/10.1145/1011767.1011781}.

\bibitem[Silver et~al.(2021)Silver, Singh, Precup, and Sutton]{SILVER2021103535}
Silver, D., Singh, S., Precup, D., and Sutton, R.~S.
\newblock Reward is enough.
\newblock \emph{Artificial Intelligence}, 299:\penalty0 103535, 2021.
\newblock ISSN 0004-3702.
\newblock \doi{https://doi.org/10.1016/j.artint.2021.103535}.
\newblock URL \url{https://www.sciencedirect.com/science/article/pii/S0004370221000862}.

\bibitem[Snell et~al.(2024)Snell, Lee, Xu, and Kumar]{snell2024scaling}
Snell, C., Lee, J., Xu, K., and Kumar, A.
\newblock Scaling llm test-time compute optimally can be more effective than scaling model parameters.
\newblock \emph{arXiv preprint arXiv:2408.03314}, 2024.

\bibitem[Stechly et~al.(2024)Stechly, Valmeekam, and Kambhampati]{stechly2024chain}
Stechly, K., Valmeekam, K., and Kambhampati, S.
\newblock Chain of thoughtlessness: An analysis of cot in planning.
\newblock \emph{arXiv preprint arXiv:2405.04776}, 2024.

\bibitem[Strobl et~al.(2024)Strobl, Merrill, Weiss, Chiang, and Angluin]{Strobl_2024}
Strobl, L., Merrill, W., Weiss, G., Chiang, D., and Angluin, D.
\newblock What formal languages can transformers express? a survey.
\newblock \emph{Transactions of the Association for Computational Linguistics}, 12:\penalty0 543–561, 2024.
\newblock ISSN 2307-387X.
\newblock \doi{10.1162/tacl_a_00663}.
\newblock URL \url{http://dx.doi.org/10.1162/tacl_a_00663}.

\bibitem[Strubell et~al.(2019)Strubell, Ganesh, and McCallum]{strubell2019energypolicyconsiderationsdeep}
Strubell, E., Ganesh, A., and McCallum, A.
\newblock Energy and policy considerations for deep learning in nlp, 2019.
\newblock URL \url{https://arxiv.org/abs/1906.02243}.

\bibitem[Sucholutsky et~al.(2024)Sucholutsky, Muttenthaler, Weller, Peng, Bobu, Kim, Love, Cueva, Grant, Groen, Achterberg, Tenenbaum, Collins, Hermann, Oktar, Greff, Hebart, Cloos, Kriegeskorte, Jacoby, Zhang, Marjieh, Geirhos, Chen, Kornblith, Rane, Konkle, O'Connell, Unterthiner, Lampinen, Müller, Toneva, and Griffiths]{sucholutsky2024gettingalignedrepresentationalalignment}
Sucholutsky, I., Muttenthaler, L., Weller, A., Peng, A., Bobu, A., Kim, B., Love, B.~C., Cueva, C.~J., Grant, E., Groen, I., Achterberg, J., Tenenbaum, J.~B., Collins, K.~M., Hermann, K.~L., Oktar, K., Greff, K., Hebart, M.~N., Cloos, N., Kriegeskorte, N., Jacoby, N., Zhang, Q., Marjieh, R., Geirhos, R., Chen, S., Kornblith, S., Rane, S., Konkle, T., O'Connell, T.~P., Unterthiner, T., Lampinen, A.~K., Müller, K.-R., Toneva, M., and Griffiths, T.~L.
\newblock Getting aligned on representational alignment, 2024.
\newblock URL \url{https://arxiv.org/abs/2310.13018}.

\bibitem[Sundararajan et~al.(2017)Sundararajan, Taly, and Yan]{sundararajan2017axiomatic}
Sundararajan, M., Taly, A., and Yan, Q.
\newblock Axiomatic attribution for deep networks.
\newblock In \emph{International conference on machine learning}, pp.\  3319--3328. PMLR, 2017.

\bibitem[Sutton(2019)]{bitterlesson2019}
Sutton, R.
\newblock The bitter lesson, 2019.
\newblock URL \url{http://www.incompleteideas.net/IncIdeas/BitterLesson.html}.

\bibitem[Swartout \& Moore(1993)Swartout and Moore]{10.1007/978-3-642-77927-5_24}
Swartout, W.~R. and Moore, J.~D.
\newblock Explanation in second generation expert systems.
\newblock In David, J.-M., Krivine, J.-P., and Simmons, R. (eds.), \emph{Second Generation Expert Systems}, pp.\  543--585, Berlin, Heidelberg, 1993. Springer Berlin Heidelberg.
\newblock ISBN 978-3-642-77927-5.

\bibitem[Syed et~al.(2024)Syed, Rager, and Conmy]{syed-etal-2024-attribution}
Syed, A., Rager, C., and Conmy, A.
\newblock Attribution patching outperforms automated circuit discovery.
\newblock In Belinkov, Y., Kim, N., Jumelet, J., Mohebbi, H., Mueller, A., and Chen, H. (eds.), \emph{Proceedings of the 7th BlackboxNLP Workshop: Analyzing and Interpreting Neural Networks for NLP}, pp.\  407--416, Miami, Florida, US, November 2024. Association for Computational Linguistics.
\newblock \doi{10.18653/v1/2024.blackboxnlp-1.25}.
\newblock URL \url{https://aclanthology.org/2024.blackboxnlp-1.25/}.

\bibitem[Talon et~al.(2024)Talon, Lippe, James, Bue, and Magliacane]{pmlr-v236-talon24a}
Talon, D., Lippe, P., James, S., Bue, A.~D., and Magliacane, S.
\newblock Towards the reusability and compositionality of causal representations.
\newblock In Locatello, F. and Didelez, V. (eds.), \emph{Proceedings of the Third Conference on Causal Learning and Reasoning}, volume 236 of \emph{Proceedings of Machine Learning Research}, pp.\  296--324. PMLR, 01--03 Apr 2024.
\newblock URL \url{https://proceedings.mlr.press/v236/talon24a.html}.

\bibitem[Tarzanagh et~al.(2023)Tarzanagh, Li, Thrampoulidis, and Oymak]{tarzanagh2023transformers}
Tarzanagh, D.~A., Li, Y., Thrampoulidis, C., and Oymak, S.
\newblock Transformers as support vector machines.
\newblock In \emph{NeurIPS 2023 Workshop on Mathematics of Modern Machine Learning}, 2023.
\newblock URL \url{https://openreview.net/forum?id=gLwzzmh79K}.

\bibitem[Templeton et~al.(2024)Templeton, Conerly, Marcus, Lindsey, Bricken, Chen, Pearce, Citro, Ameisen, Jones, Cunningham, Turner, McDougall, MacDiarmid, Freeman, Sumers, Rees, Batson, Jermyn, Carter, Olah, and Henighan]{templeton2024scaling}
Templeton, A., Conerly, T., Marcus, J., Lindsey, J., Bricken, T., Chen, B., Pearce, A., Citro, C., Ameisen, E., Jones, A., Cunningham, H., Turner, N.~L., McDougall, C., MacDiarmid, M., Freeman, C.~D., Sumers, T.~R., Rees, E., Batson, J., Jermyn, A., Carter, S., Olah, C., and Henighan, T.
\newblock Scaling monosemanticity: Extracting interpretable features from claude 3 sonnet.
\newblock \emph{Transformer Circuits Thread}, 2024.
\newblock URL \url{https://transformer-circuits.pub/2024/scaling-monosemanticity/index.html}.

\bibitem[Tigges et~al.(2024)Tigges, Hanna, Yu, and Biderman]{tigges2024llmcircuitanalysesconsistent}
Tigges, C., Hanna, M., Yu, Q., and Biderman, S.
\newblock Llm circuit analyses are consistent across training and scale, 2024.
\newblock URL \url{https://arxiv.org/abs/2407.10827}.

\bibitem[Todd et~al.(2024)Todd, Li, Sharma, Mueller, Wallace, and Bau]{todd2024function}
Todd, E., Li, M., Sharma, A.~S., Mueller, A., Wallace, B.~C., and Bau, D.
\newblock Function vectors in large language models.
\newblock In \emph{The Twelfth International Conference on Learning Representations}, 2024.
\newblock URL \url{https://openreview.net/forum?id=AwyxtyMwaG}.

\bibitem[Tsai et~al.(2019)Tsai, Bai, Yamada, Morency, and Salakhutdinov]{tsai-etal-2019-transformer}
Tsai, Y.-H.~H., Bai, S., Yamada, M., Morency, L.-P., and Salakhutdinov, R.
\newblock Transformer dissection: An unified understanding for transformer`s attention via the lens of kernel.
\newblock In Inui, K., Jiang, J., Ng, V., and Wan, X. (eds.), \emph{Proceedings of the 2019 Conference on Empirical Methods in Natural Language Processing and the 9th International Joint Conference on Natural Language Processing (EMNLP-IJCNLP)}, pp.\  4344--4353, Hong Kong, China, November 2019. Association for Computational Linguistics.
\newblock \doi{10.18653/v1/D19-1443}.
\newblock URL \url{https://aclanthology.org/D19-1443/}.

\bibitem[Turing(1936)]{turing1936a}
Turing, A.~M.
\newblock On computable numbers, with an application to the {E}ntscheidungsproblem.
\newblock \emph{Proceedings of the London Mathematical Society}, 2\penalty0 (42):\penalty0 230--265, 1936.
\newblock URL \url{http://www.cs.helsinki.fi/u/gionis/cc05/OnComputableNumbers.pdf}.

\bibitem[Turpin et~al.(2024)Turpin, Michael, Perez, and Bowman]{turpin2024language}
Turpin, M., Michael, J., Perez, E., and Bowman, S.
\newblock Language models don't always say what they think: unfaithful explanations in chain-of-thought prompting.
\newblock \emph{Advances in Neural Information Processing Systems}, 36, 2024.

\bibitem[Vasileiou \& Eberle(2024)Vasileiou and Eberle]{vasileiou-2024-explaining}
Vasileiou, A. and Eberle, O.
\newblock Explaining text similarity in transformer models.
\newblock In \emph{Proceedings of the 2024 Conference of the North American Chapter of the Association for Computational Linguistics: Human Language Technologies (Volume 1: Long Papers)}, pp.\  7859--7873, Mexico City, Mexico, June 2024. Association for Computational Linguistics.
\newblock \doi{10.18653/v1/2024.naacl-long.435}.
\newblock URL \url{https://aclanthology.org/2024.naacl-long.435/}.

\bibitem[Vaswani(2017)]{vaswani2017attention}
Vaswani, A.
\newblock Attention is all you need.
\newblock \emph{Advances in Neural Information Processing Systems}, 2017.

\bibitem[Veli{\v{c}}kovi{\'c} \& Blundell(2021)Veli{\v{c}}kovi{\'c} and Blundell]{velivckovic2021neural}
Veli{\v{c}}kovi{\'c}, P. and Blundell, C.
\newblock Neural algorithmic reasoning.
\newblock \emph{Patterns}, 2\penalty0 (7), 2021.

\bibitem[Vig \& Belinkov(2019)Vig and Belinkov]{vig2019analyzingstructureattentiontransformer}
Vig, J. and Belinkov, Y.
\newblock Analyzing the structure of attention in a transformer language model, 2019.
\newblock URL \url{https://arxiv.org/abs/1906.04284}.

\bibitem[Vilas et~al.(2024)Vilas, Adolfi, Poeppel, and Roig]{pmlr-v235-vilas24a}
Vilas, M.~G., Adolfi, F., Poeppel, D., and Roig, G.
\newblock Position: An inner interpretability framework for {AI} inspired by lessons from cognitive neuroscience.
\newblock In Salakhutdinov, R., Kolter, Z., Heller, K., Weller, A., Oliver, N., Scarlett, J., and Berkenkamp, F. (eds.), \emph{Proceedings of the 41st International Conference on Machine Learning}, volume 235 of \emph{Proceedings of Machine Learning Research}, pp.\  49506--49522. PMLR, 21--27 Jul 2024.
\newblock URL \url{https://proceedings.mlr.press/v235/vilas24a.html}.

\bibitem[Villalobos et~al.(2024)Villalobos, Ho, Sevilla, Besiroglu, Heim, and Hobbhahn]{10.5555/3692070.3694094}
Villalobos, P., Ho, A., Sevilla, J., Besiroglu, T., Heim, L., and Hobbhahn, M.
\newblock Position: will we run out of data? limits of llm scaling based on human-generated data.
\newblock In \emph{Proceedings of the 41st International Conference on Machine Learning}, ICML'24. JMLR.org, 2024.

\bibitem[von Oswald et~al.(2024)von Oswald, Schlegel, Meulemans, Kobayashi, Niklasson, Zucchet, Scherrer, Miller, Sandler, y~Arcas, Vladymyrov, Pascanu, and Sacramento]{vonoswald2024uncoveringmesaoptimizationalgorithmstransformers}
von Oswald, J., Schlegel, M., Meulemans, A., Kobayashi, S., Niklasson, E., Zucchet, N., Scherrer, N., Miller, N., Sandler, M., y~Arcas, B.~A., Vladymyrov, M., Pascanu, R., and Sacramento, J.
\newblock Uncovering mesa-optimization algorithms in transformers, 2024.
\newblock URL \url{https://arxiv.org/abs/2309.05858}.

\bibitem[Wang et~al.(2023)Wang, Variengien, Conmy, Shlegeris, and Steinhardt]{wang2023interpretability}
Wang, K.~R., Variengien, A., Conmy, A., Shlegeris, B., and Steinhardt, J.
\newblock Interpretability in the wild: a circuit for indirect object identification in {GPT}-2 small.
\newblock In \emph{The Eleventh International Conference on Learning Representations}, 2023.
\newblock URL \url{https://openreview.net/forum?id=NpsVSN6o4ul}.

\bibitem[Webb et~al.(2024)Webb, Mondal, and Momennejad]{webb2024LLMPFC}
Webb, T., Mondal, S.~S., and Momennejad, I.
\newblock Improving planning with large language models: A modular agentic architecture, 2024.
\newblock URL \url{https://arxiv.org/abs/2310.00194}.

\bibitem[Wei et~al.(2022)Wei, Wang, Schuurmans, Bosma, Xia, Chi, Le, Zhou, et~al.]{wei2022chain}
Wei, J., Wang, X., Schuurmans, D., Bosma, M., Xia, F., Chi, E., Le, Q.~V., Zhou, D., et~al.
\newblock Chain-of-thought prompting elicits reasoning in large language models.
\newblock \emph{Advances in neural information processing systems}, 35:\penalty0 24824--24837, 2022.

\bibitem[Weiss et~al.(2021)Weiss, Goldberg, and Yahav]{pmlr-v139-weiss21a}
Weiss, G., Goldberg, Y., and Yahav, E.
\newblock Thinking like transformers.
\newblock In Meila, M. and Zhang, T. (eds.), \emph{Proceedings of the 38th International Conference on Machine Learning}, volume 139 of \emph{Proceedings of Machine Learning Research}, pp.\  11080--11090. PMLR, 18--24 Jul 2021.
\newblock URL \url{https://proceedings.mlr.press/v139/weiss21a.html}.

\bibitem[Wiedemer et~al.(2023)Wiedemer, Mayilvahanan, Bethge, and Brendel]{wiedemer2023compositional}
Wiedemer, T., Mayilvahanan, P., Bethge, M., and Brendel, W.
\newblock Compositional generalization from first principles.
\newblock In \emph{Thirty-seventh Conference on Neural Information Processing Systems}, 2023.
\newblock URL \url{https://openreview.net/forum?id=LqOQ1uJmSx}.

\bibitem[Wiegreffe \& Pinter(2019)Wiegreffe and Pinter]{wiegreffe-pinter-2019-attention}
Wiegreffe, S. and Pinter, Y.
\newblock Attention is not not explanation.
\newblock In Inui, K., Jiang, J., Ng, V., and Wan, X. (eds.), \emph{Proceedings of the 2019 Conference on Empirical Methods in Natural Language Processing and the 9th International Joint Conference on Natural Language Processing (EMNLP-IJCNLP)}, pp.\  11--20, Hong Kong, China, November 2019. Association for Computational Linguistics.
\newblock \doi{10.18653/v1/D19-1002}.
\newblock URL \url{https://aclanthology.org/D19-1002/}.

\bibitem[Williams et~al.(2021)Williams, Kunz, Kornblith, and Linderman]{Pro_NEURIPS2021_252a3dba}
Williams, A.~H., Kunz, E., Kornblith, S., and Linderman, S.
\newblock Generalized shape metrics on neural representations.
\newblock In Ranzato, M., Beygelzimer, A., Dauphin, Y., Liang, P., and Vaughan, J.~W. (eds.), \emph{Advances in Neural Information Processing Systems}, volume~34, pp.\  4738--4750. Curran Associates, Inc., 2021.
\newblock URL \url{https://proceedings.neurips.cc/paper_files/paper/2021/file/252a3dbaeb32e7690242ad3b556e626b-Paper.pdf}.

\bibitem[Wu et~al.(2023{\natexlab{a}})Wu, Bansal, Zhang, Wu, Li, Zhu, Jiang, Zhang, Zhang, Liu, Awadallah, White, Burger, and Wang]{wu2023autogen}
Wu, Q., Bansal, G., Zhang, J., Wu, Y., Li, B., Zhu, E., Jiang, L., Zhang, X., Zhang, S., Liu, J., Awadallah, A.~H., White, R.~W., Burger, D., and Wang, C.
\newblock Autogen: Enabling next-gen llm applications via multi-agent conversation, 2023{\natexlab{a}}.
\newblock URL \url{https://arxiv.org/abs/2308.08155}.

\bibitem[Wu et~al.(2023{\natexlab{b}})Wu, Geiger, Icard, Potts, and Goodman]{NEURIPS2023_f6a8b109}
Wu, Z., Geiger, A., Icard, T., Potts, C., and Goodman, N.
\newblock Interpretability at scale: Identifying causal mechanisms in alpaca.
\newblock In Oh, A., Naumann, T., Globerson, A., Saenko, K., Hardt, M., and Levine, S. (eds.), \emph{Advances in Neural Information Processing Systems}, volume~36, pp.\  78205--78226. Curran Associates, Inc., 2023{\natexlab{b}}.

\bibitem[Yang et~al.(2024)Yang, Sun, and Buzsaki]{yang2024interpretability}
Yang, W., Sun, C., and Buzsaki, G.
\newblock {INTERPRETABILITY} {OF} {LLM} {DECEPTION}: {UNIVERSAL} {MOTIF}.
\newblock In \emph{Neurips Safe Generative AI Workshop 2024}, 2024.
\newblock URL \url{https://openreview.net/forum?id=DRWCDFsb2e}.

\bibitem[Yao et~al.(2024)Yao, Yu, Zhao, Shafran, Griffiths, Cao, and Narasimhan]{yao2024tree}
Yao, S., Yu, D., Zhao, J., Shafran, I., Griffiths, T., Cao, Y., and Narasimhan, K.
\newblock Tree of thoughts: Deliberate problem solving with large language models.
\newblock \emph{Advances in Neural Information Processing Systems}, 36, 2024.

\bibitem[Ye et~al.(2025)Ye, Xu, Li, and {Allen-Zhu}]{YXLA2024-gsm1}
Ye, T., Xu, Z., Li, Y., and {Allen-Zhu}, Z.
\newblock {Physics of Language Models: Part 2.1, Grade-School Math and the Hidden Reasoning Process}.
\newblock In \emph{Proceedings of the 13th International Conference on Learning Representations}, ICLR~'25, April 2025.
\newblock Full version available at \url{http://arxiv.org/abs/2407.20311}.

\bibitem[Yousefi et~al.(2024)Yousefi, Betthauser, Hasanbeig, Millière, and Momennejad]{yousefi2024decodingincontextlearningneuroscienceinspired}
Yousefi, S., Betthauser, L., Hasanbeig, H., Millière, R., and Momennejad, I.
\newblock Decoding in-context learning: Neuroscience-inspired analysis of representations in large language models, 2024.
\newblock URL \url{https://arxiv.org/abs/2310.00313}.

\bibitem[Yu et~al.(2023)Yu, Merullo, and Pavlick]{yu-etal-2023-characterizing}
Yu, Q., Merullo, J., and Pavlick, E.
\newblock Characterizing mechanisms for factual recall in language models.
\newblock In \emph{Proceedings of the 2023 Conference on Empirical Methods in Natural Language Processing}, pp.\  9924--9959, Singapore, December 2023. Association for Computational Linguistics.
\newblock \doi{10.18653/v1/2023.emnlp-main.615}.
\newblock URL \url{https://aclanthology.org/2023.emnlp-main.615/}.

\bibitem[Zekri et~al.(2024)Zekri, Odonnat, Benechehab, Bleistein, Boullé, and Redko]{zekri2024largelanguagemodelsmarkov}
Zekri, O., Odonnat, A., Benechehab, A., Bleistein, L., Boullé, N., and Redko, I.
\newblock Large language models as markov chains, 2024.
\newblock URL \url{https://arxiv.org/abs/2410.02724}.

\bibitem[Zhong et~al.(2023)Zhong, Liu, Tegmark, and Andreas]{zhong2023the}
Zhong, Z., Liu, Z., Tegmark, M., and Andreas, J.
\newblock The clock and the pizza: Two stories in mechanistic explanation of neural networks.
\newblock In \emph{Thirty-seventh Conference on Neural Information Processing Systems}, 2023.
\newblock URL \url{https://openreview.net/forum?id=S5wmbQc1We}.

\bibitem[Zhou et~al.(2018)Zhou, Bau, Oliva, and Torralba]{zhou2018interpreting}
Zhou, B., Bau, D., Oliva, A., and Torralba, A.
\newblock Interpreting deep visual representations via network dissection.
\newblock \emph{IEEE transactions on pattern analysis and machine intelligence}, 41\penalty0 (9):\penalty0 2131--2145, 2018.

\bibitem[Zhou et~al.(2024)Zhou, Bradley, Littwin, Razin, Saremi, Susskind, Bengio, and Nakkiran]{zhou2024what}
Zhou, H., Bradley, A., Littwin, E., Razin, N., Saremi, O., Susskind, J.~M., Bengio, S., and Nakkiran, P.
\newblock What algorithms can transformers learn? a study in length generalization.
\newblock In \emph{The Twelfth International Conference on Learning Representations}, 2024.
\newblock URL \url{https://openreview.net/forum?id=AssIuHnmHX}.

\bibitem[Zou et~al.(2023)Zou, Phan, Chen, Campbell, Guo, Ren, Pan, Yin, Mazeika, Dombrowski, Goel, Li, Byun, Wang, Mallen, Basart, Koyejo, Song, Fredrikson, Kolter, and Hendrycks]{zou2023representationengineeringtopdownapproach}
Zou, A., Phan, L., Chen, S., Campbell, J., Guo, P., Ren, R., Pan, A., Yin, X., Mazeika, M., Dombrowski, A.-K., Goel, S., Li, N., Byun, M.~J., Wang, Z., Mallen, A., Basart, S., Koyejo, S., Song, D., Fredrikson, M., Kolter, J.~Z., and Hendrycks, D.
\newblock Representation engineering: A top-down approach to ai transparency, 2023.
\newblock URL \url{https://arxiv.org/abs/2310.01405}.

\end{thebibliography}
\bibliographystyle{icml2025}

\newpage
\appendix
\onecolumn
\section{Appendix}

\subsection{Statistical Testing: Paired-sample \textit{t}-tests for attention from the final token to correct vs. incorrect pathways.} 

Table \ref{app:tab:ttest_attn_correct}: Reports statistics for the linear mixed-effects model, testing attention directed from the final token to the correct versus incorrect pathway across all layers. Mixed-effects modeling was conducted using the lmerTest package in R. 

Table \ref{app:tab:ttest_attn_correct_layers}: Reports specific layers with significantly greater attention to the correct pathway, while Tab. \ref{app:tab:ttest_attn_incorrect_layers} reports layers with significantly greater attention to the incorrect pathway. Note that conducting multiple t-tests (one for each layer) increases the risk of Type I errors (false positives). With $k$ layers, we have $k$ chances to incorrectly reject the null hypothesis. To mitigate the possibility of Type I errors, we used Bonferroni to control the family-wise error rate.

\begin{table}[h!]
    \scriptsize
    \centering
    \setlength{\tabcolsep}{5pt}
    \caption{Greater attention to the correct pathway across layers.}
    \label{app:tab:ttest_attn_correct}
    \vskip 5pt
    \begin{tabular}{l c c c c}
        \toprule
        \textbf{\textit{b}} &
        \textbf{\textit{SE}} &
        \textbf{\textit{t}-statistic} & 
        \textbf{\textit{df}} 
        & \textbf{\textit{p}-value } \\
        \midrule
        0.33 & 0.07 & 4.51 & 2015 & $p=7.0\!\times\!10^{-6}$ \\
        \bottomrule
    \end{tabular}
\end{table}

\vspace{-.1cm}
\begin{table}[h!]
    \scriptsize
    \centering
    \setlength{\tabcolsep}{11pt}
        \caption{Greater attention to the correct pathway (individual layers).}
    \label{app:tab:ttest_attn_correct_layers}
    \vskip 5pt
    \begin{tabular}{l c c l}
        \toprule
        \textbf{Layer} & \textbf{\textit{t}-Statistic} & \textbf{\textit{df}} & \textbf{\textit{p}-Value} \\
        \midrule
        6 & 5.68 & 31 & $\textit{p} = 1.5\!\times\!10^{-6}$ \\
        7 & 12.18 & 31 & $\textit{p} = 1.17\!\times\!10^{-13}$ \\
        10 & 9.47 & 31 & $\textit{p} = 5.86\!\times\!10^{-11}$ \\
        11 & 8.97 & 31 & $\textit{p} = 1.99\!\times\!10^{-10}$ \\
        12 & 4.32 & 31 & $\textit{p} = 7.49\!\times\!10^{-5}$ \\
        13 & 7.13 & 31 & $\textit{p} = 2.60\!\times\!10^{-8}$ \\
        14 & 3.84 & 31 & $\textit{p} = 2.87\!\times\!10^{-4}$ \\
        16 & 4.36 & 31 & $\textit{p} = 6.65\!\times\!10^{-5}$ \\
        17 & 4.60 & 31 & $\textit{p} = 3.40\!\times\!10^{-5}$ \\
        20 & 3.71 & 31 & $\textit{p} = 4.1\!\times\!10^{-4}$ \\
        21 & 4.73 & 31 & $\textit{p} = 2.36\!\times\!10^{-5}$ \\
        26 & 3.38 & 31 & $\textit{p} = 9.9\!\times\!10^{-4}$ \\
        29 & 5.96 & 31 & $\textit{p} = 6.97\!\times\!10^{-7}$ \\
        30 & 4.00 & 31 & $\textit{p} = 1.83\!\times\!10^{-4}$ \\
        \bottomrule
    \end{tabular}
\end{table}

\vspace{-.1cm}
\begin{table}[h!]
    \scriptsize
    \centering
    \setlength{\tabcolsep}{9pt}
    \caption{Greater attention to the incorrect pathway (individual layers).}
    \label{app:tab:ttest_attn_incorrect_layers}
    \vskip 5pt
    
    \begin{tabular}{l c c c}
        \toprule
        \textbf{Layer} & \textbf{\textit{t}-statistic} & \textbf{\textit{df}} & \textbf{\textit{p}-value } \\
        \midrule
        0 & -5.34 & 31 & $\textit{p} = 4.05\!\times\!10^{-6}$ \\
        2 & -4.58 & 31 & $\textit{p} = 3.52\!\times\!10^{-5}$ \\
        23 & -3.23 & 31 & $\textit{p} = 1.47\!\times\!10^{-3}$ \\

        \bottomrule
    \end{tabular}
\end{table}

\newpage

\subsection{Attention from the Goal Location to all Nodes for the Tree Graph (n=7)}
\textbf{Prompts}

(1) From the hotel lobby, there are various rooms denoted by letter names. The lobby connects to O, which connects to W and Q. The lobby also connects to G, which connects to V and M. How can someone get to \textbf{W/Q/V/M} from the lobby?

\begin{figure}[H]
\begin{center}
    \centering
    \includegraphics[width=0.7\columnwidth]{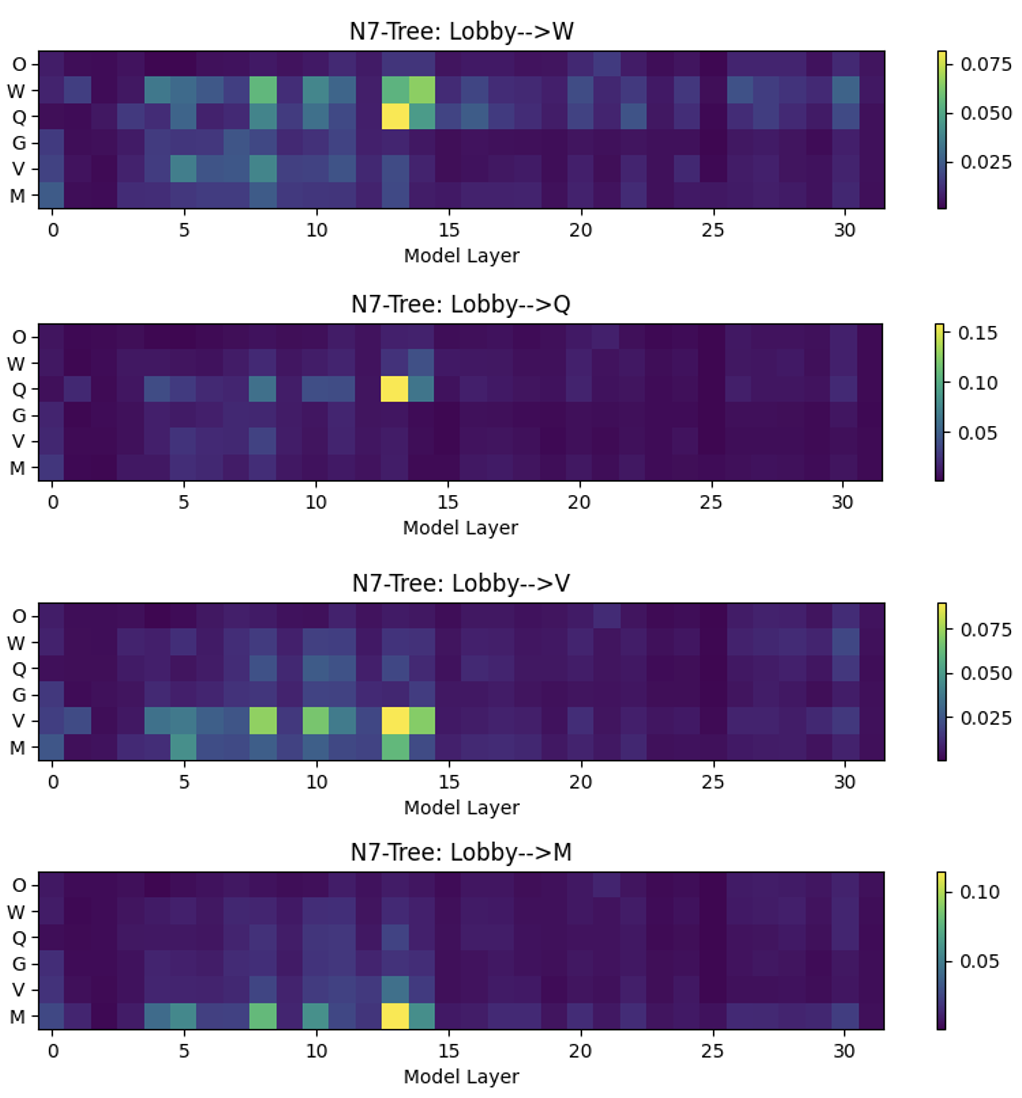}
    \caption{Attention heatmaps from the goal token to all graph nodes in the tree graph, when each final node is specified as the goal location.}
    \end{center}
    \vskip -0.2in
    \label{fig:n7tree_AllPaths}
\end{figure}

\newpage 
\subsection{Analyzing Representations} \label{sec:rep_appendix}
\subsection*{Prompts }
(1) Given a description of a set of rooms, determine if someone can get to W from the lobby? 
Only answer "Yes" or "No". 
The description is: From the hotel lobby, there are various rooms denoted by letter names. The lobby connects to O, which connects to W and Q. The lobby also connects to G, which connects to V and M. 
Answer: 

(2) Given a description of a set of rooms, determine if someone can get to Q from the lobby? 
Only answer "Yes" or "No". 
The description is: From the hotel lobby, there are various rooms denoted by letter names. The lobby connects to O, which connects to W and Q. The lobby also connects to G, which connects to V and M. 
Answer: 

(3) Given a description of a set of rooms, determine if someone can get to V from the lobby? 
Only answer "Yes" or "No". 
The description is: From the hotel lobby, there are various rooms denoted by letter names. The lobby connects to O, which connects to W and Q. The lobby also connects to G, which connects to V and M. 
Answer: 

(4) Given a description of a set of rooms, determine if someone can get to M from the lobby? 
Only answer "Yes" or "No". 
The description is: From the hotel lobby, there are various rooms denoted by letter names. The lobby connects to O, which connects to W and Q. The lobby also connects to G, which connects to V and M. 
Answer: 

\begin{figure*}[htbp]
    \centering
    \begin{tabular}{cc}
        \includegraphics[width=0.5\columnwidth]{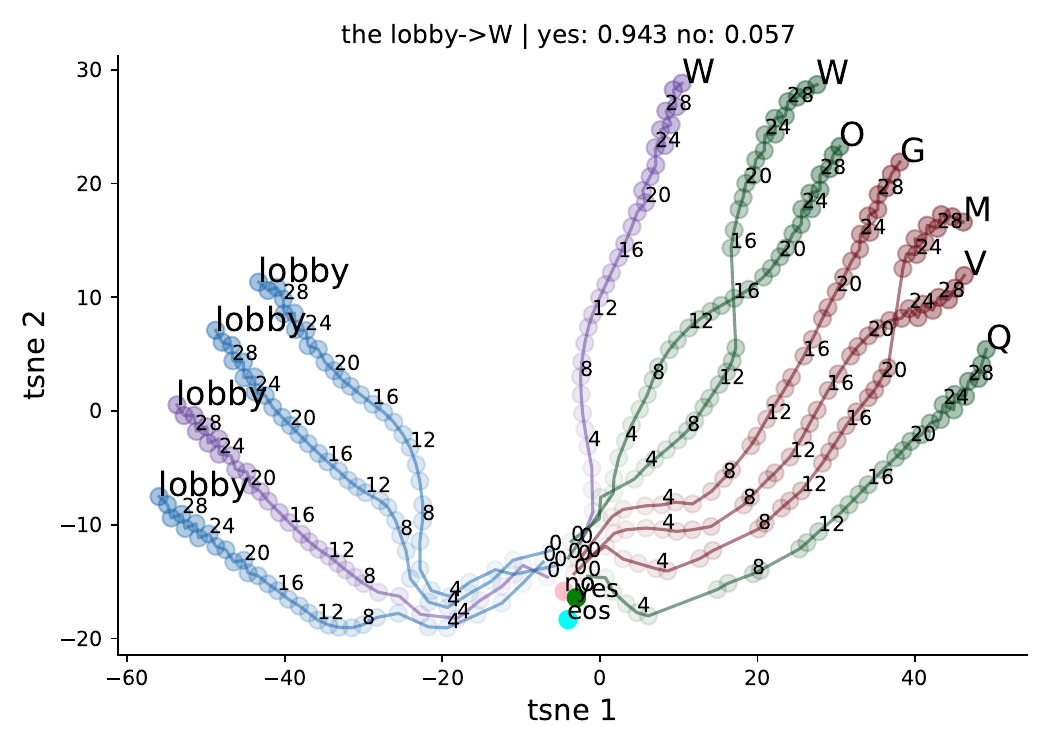} &  
        \includegraphics[width=0.5\columnwidth]{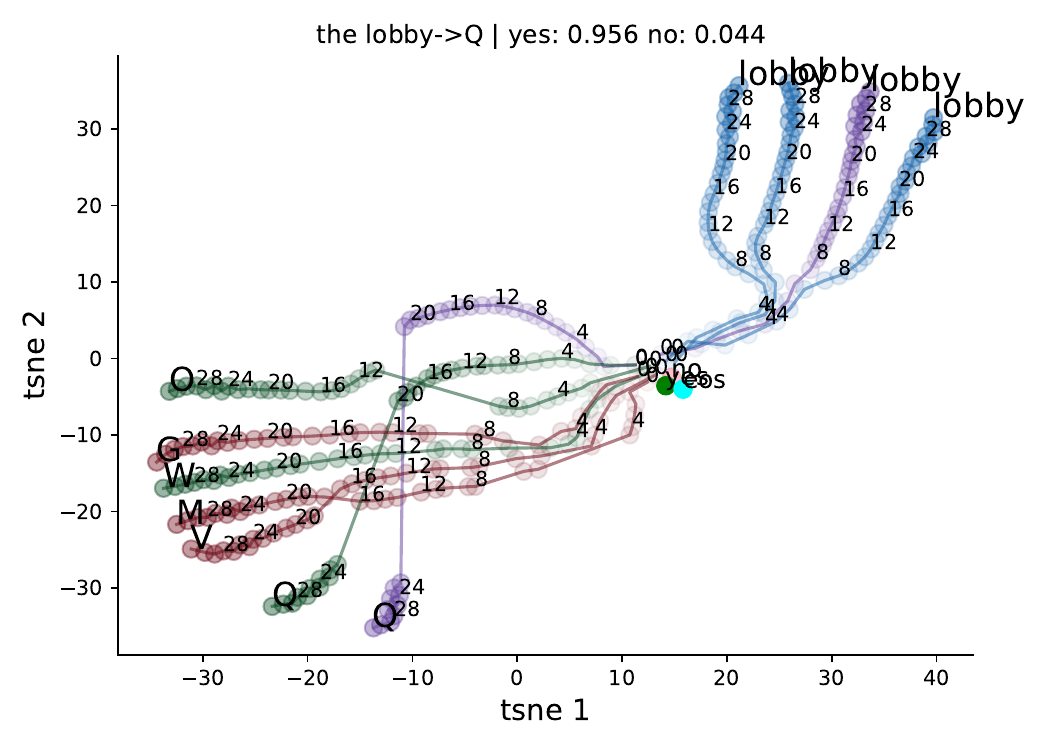}  \\ 
        (1) Start from Lobby: 2-2, Goal W & (2) Start from Lobby: 2-2, Goal Q \\[6pt]
        \includegraphics[width=0.5\columnwidth]{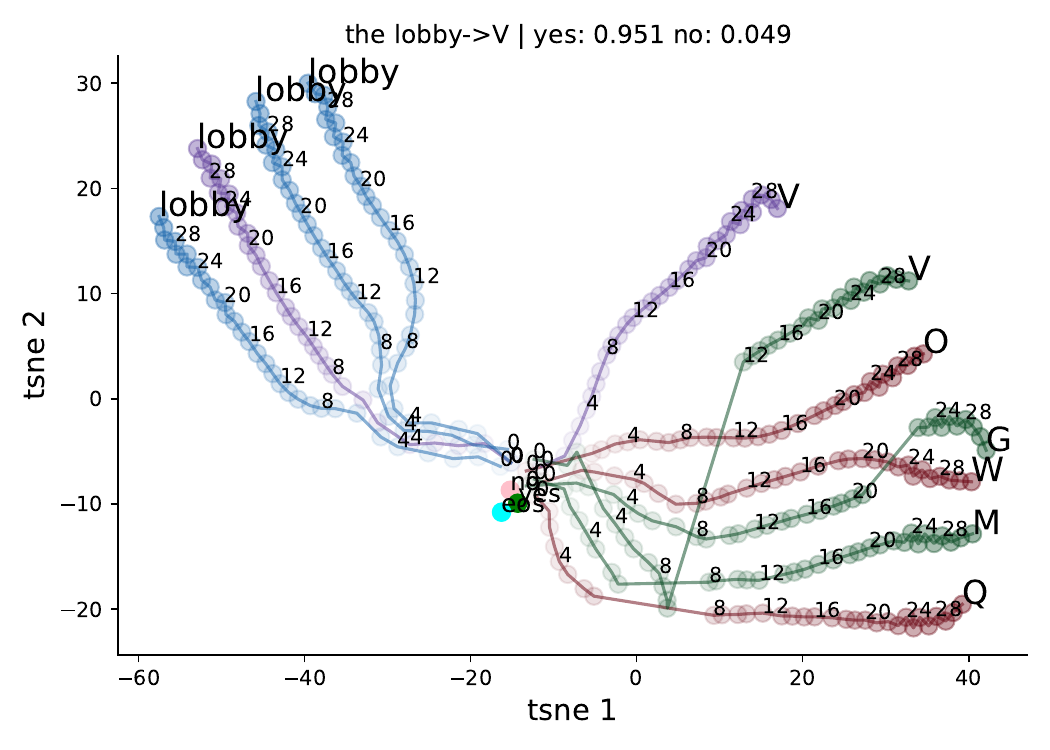} &  
        \includegraphics[width=0.5\columnwidth]{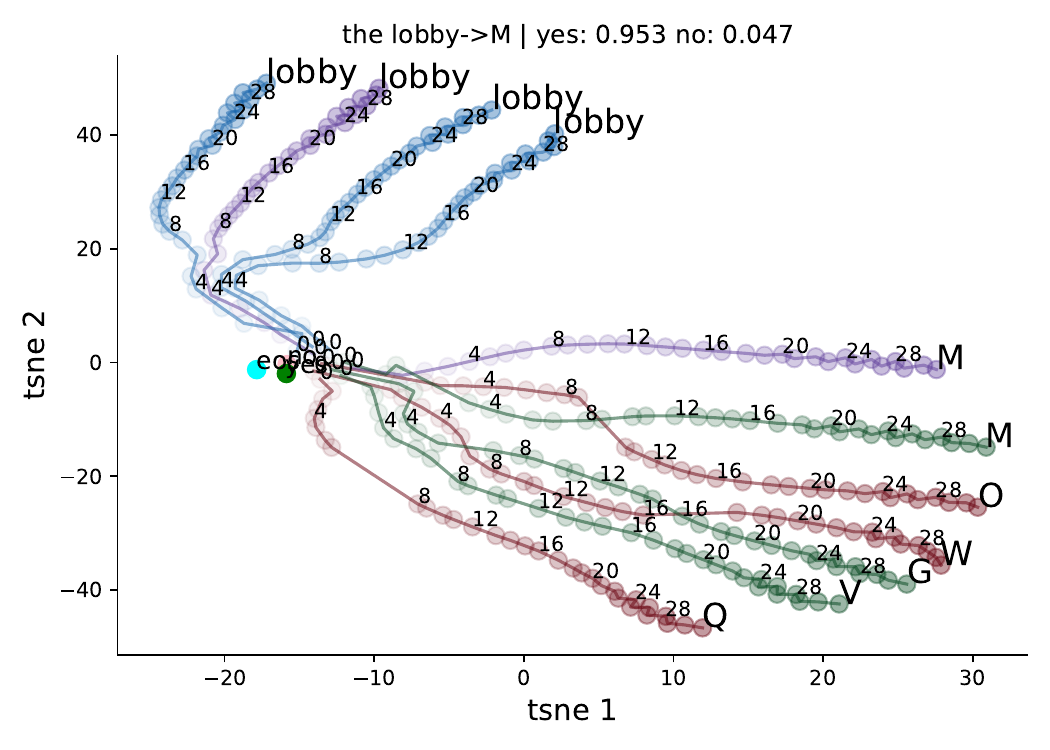} \\ 
        (3) Start from Lobby: 2-2, Goal V & (4) Start from Lobby: 2-2, Goal M \\[6pt]
    \end{tabular}
    \caption{T-SNE projections of activations in a graph search setting. Plots correspond to prompts (1)--(4), corresponding to a changing goal node in a tree graph of n=7 nodes (see Figure~ \ref{fig:caseSchema}). The probabilities of generating the correct `yes' token as compared to the `no' token using the final token representation is given for each setting.}
    \label{fig:tsne2x2}
\end{figure*}

\clearpage
\begin{figure}
    \centering
    \includegraphics[width=1\linewidth]{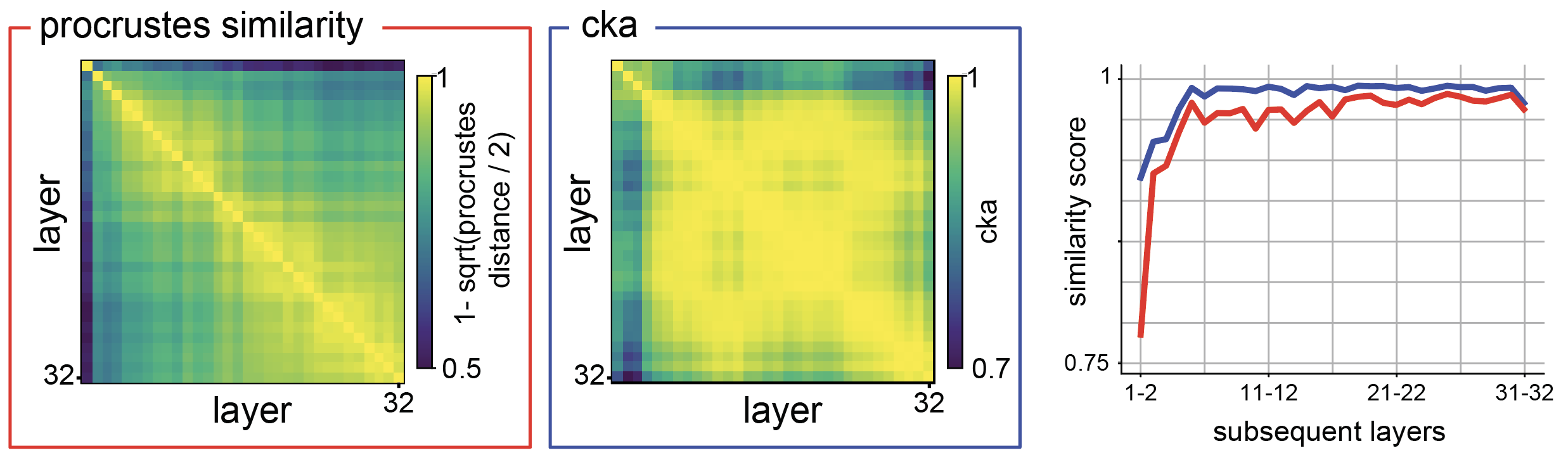}
    \caption{\textit{Left:} We reasoned that discrete steps in a BFS or DFS rollout might be identified as substantial changes in representational geometry between subsequent layers.  To explore this possibility, we computed representational similarity matrices using two similarity measures: the procrustes similarity 
    \cite{Pro_NEURIPS2021_252a3dba} and the centered kernel alignment \cite{kornblith2019similarity}.    \textit{Right:} similarity scores across subsequent layers.
    }
    \label{fig:rsa_si}
\end{figure}

\subsection{Results on Llama3.1-70B-Instruct} \label{app:70b}

\begin{figure}[ht]
  \centering
  \begin{minipage}{0.5\linewidth}
    \centering
    \includegraphics[width=\linewidth]{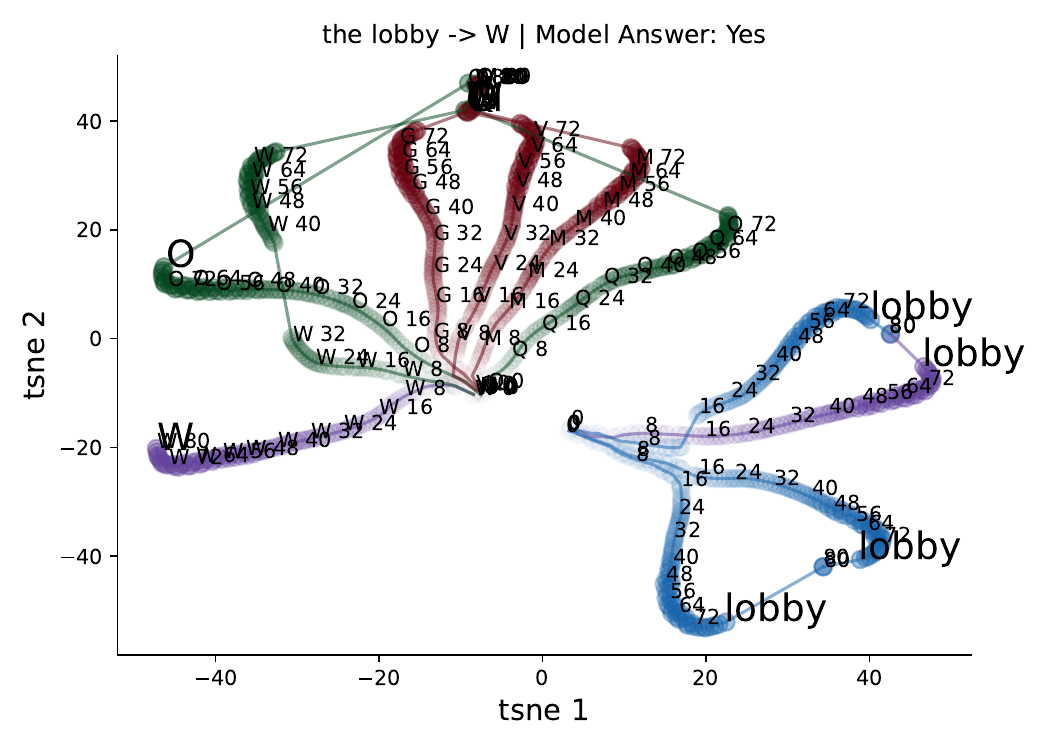}
    \vspace{-16pt}
     \caption{Representational analysis using  Llama3.1-70B. Our t-SNE results show a clear separation of room representations across layers, in particular, a separation of non-goal nodes (green  \& red) vs. the goal node (`W' in purple). See also Figure~\ref{fig:results} in the main paper for details.}
    \label{fig:tsne_70B}
  \end{minipage}
  \hfill
  \begin{minipage}{0.46\linewidth}
    \centering
    \includegraphics[width=\linewidth]{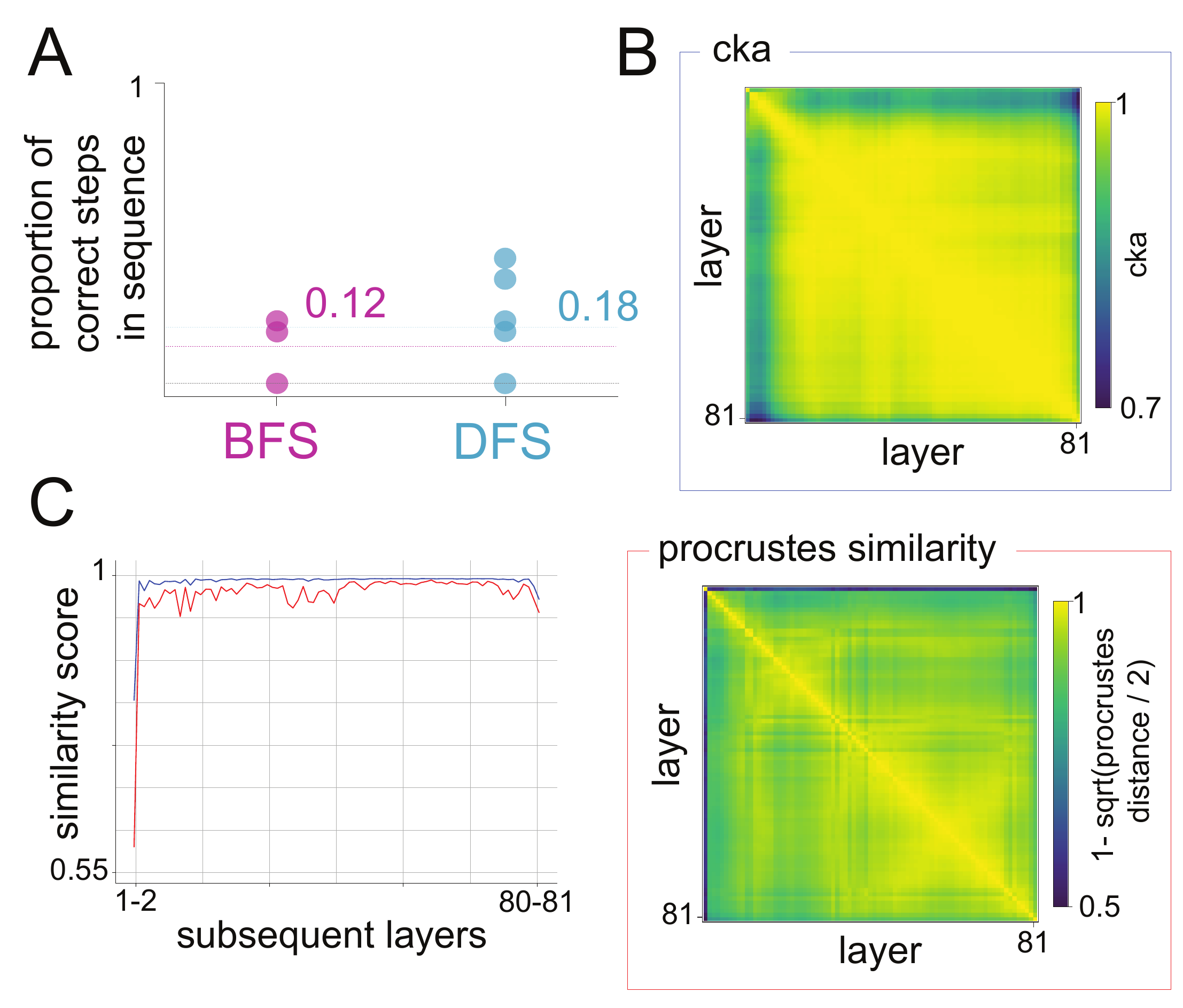}
     \caption{Evaluation of breadth-first search (BFS) and depth-first search (DFS) hypotheses in Llama3.1-70B.  (\textbf{A}) Proportion of correct steps identified in the layer by layer hidden activations. Each data point represents a single rollout of the BFS or DFS algorithm.  (\textbf{B}) Representational  similarities between layers computed using two similarity measures.  (\textbf{C}) Representational similarity between subsequent layers (off diagonal).}
    \label{fig:bfs_dfs_70B}
  \end{minipage}
\end{figure}

\begin{figure}[ht!]
  \centering
  \begin{minipage}{0.49\linewidth}
    \centering
    \includegraphics[width=\linewidth]{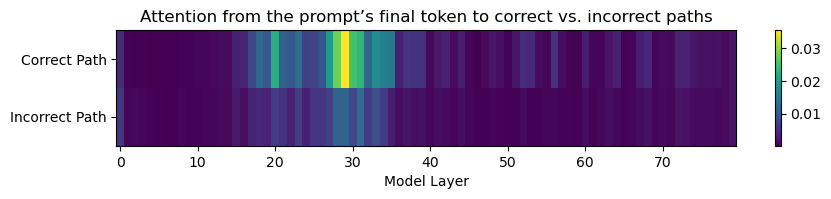}
  \end{minipage}
  \hfill
  \begin{minipage}{0.49\linewidth}
    \centering
    \includegraphics[width=\linewidth]{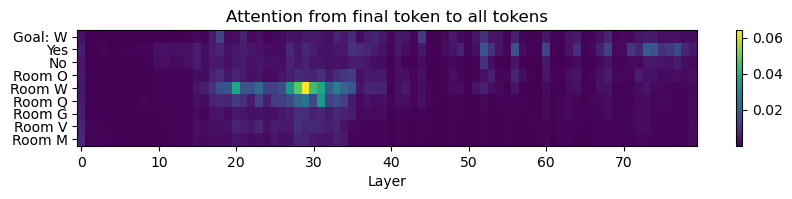}
  \end{minipage}
  \caption{Attention heatmaps for Llama-3.1-70B. \textit{Left}: Average attention per layer directed from the final token to the correct vs. incorrect pathways when node W is specified as the goal location. \textit{Right}: Average attention from final token to different graph node and response tokens.}
  \label{fig:llama70b_attention_combined}
\end{figure}

\end{document}